\documentclass[letterpaper, 10 pt, journal, twoside]{IEEEtran}

\usepackage{amsmath} % assumes amsmath package installed
\usepackage{amssymb}  % assumes amsmath package installed
\usepackage{cite}
\usepackage{physics}
\usepackage{bm}      % for bold math symbols
\usepackage{tabularx}
\usepackage{graphicx}
\usepackage{svg}
\usepackage{placeins} % For the \FloatBarrier command
\usepackage{float} % For the H specifier
\usepackage{caption}
\usepackage{subfigure}
\usepackage{array}
\usepackage{siunitx}

\makeatletter
\let\NAT@parse\undefined
\makeatother
\usepackage{hyperref}
\usepackage{fancyhdr}
\newcommand{\prths}[1]{\left(#1\right)}

\newcommand{\vecg}{\mathbf{g}}

\newcommand{\vecq}{\mathbf{q}}

\newcommand{\vecx}{\mathbf{x}}

\newcommand{\vecu}{\mathbf{u}}

\newcommand{\vecxi}{\bm{\xi}}

\newcommand{\matM}{\mathbf{M}}

\newcommand{\matR}{\bm{\mathit{R}}}
\newcommand{\matU}{\mathbf{U}}

\newcommand{\matX}{\mathbf{X}}

\newcommand{\inertia}{\mathbf{J}}

\newcommand{\loadmass}{m_{L}}
\newcommand{\half}{\frac{1}{2}}

\newcommand{\angvel}{\mathbf{\Omega}}
\newcommand{\angacc}{\dot{\angvel}}

\newcommand{\robotpos}{\vecx_{Q}}
\newcommand{\robotrot}{\matR}

\newcommand{\robotvel}{\dot{\vecx}_{Q}}
\newcommand{\robotacc}{\ddot{\vecx}_{Q}}
\newcommand{\robotangvel}{\angvel}
\newcommand{\robotangacc}{\angacc}

\newcommand{\loadpos}{\vecx_{L}}
\newcommand{\loadposdes}{\vecx_{L,des}}

\newcommand{\loadvel}{\dot{\vecx}_{L}}
\newcommand{\loadveldes}{\dot{\vecx}_{L,des}}
\newcommand{\loadacc}{\ddot{\vecx}_{L}}
\newcommand{\loadaccdes}{\ddot{\vecx}_{L,des}}

\newcommand{\cablevec}[1]{\vecxi_{#1}}

\newcommand{\cabledotvec}[1]{\dot{\vecxi}_{#1}}

\newcommand{\cableddotvec}[1]{\ddot{\vecxi}_{#1}}

\newcommand{\realnum}[1]{\mathbb{R}^{#1}}
\newcommand{\SOthree}{SO(3)}

\newcommand{\worldf}{\mathcal{I}}
\newcommand{\robotf}{\mathcal{B}}
\newcommand{\loadf}{\mathcal{L}}
\newcommand{\axis}[2]{\mathbf{e}_{#1}^{#2}}
\newcommand{\cameraf}{\mathcal{C}}
\title{
HPA-MPC: Hybrid Perception-Aware Nonlinear Model Predictive Control for Quadrotors with Suspended Loads
}

\author{Mrunal Sarvaiya$^{1}$, Guanrui Li$^{2}$, and Giuseppe Loianno$^{1}$% <-this % stops a space
\thanks{
Accepted to IEEE Robotics and Automation Letters. $^1$The authors are with the New York University, Tandon School of Engineering, Brooklyn, NY 11201, USA. {\tt\footnotesize email: \{mrunal.s, loiannog\}@nyu.edu}.}
\thanks{$^2$The author is with the Worcester Polytechnic Institute, Robotics Engineering, Worcester, MA 01609, USA. {\tt\footnotesize email: \{gli7\}@wpi.edu}.}
}
\begin{document}

%%%%%%%%%%%%%%%%%%%%%%%%%%%%%%%%%%%%%%%%%%%%%%%%%%%%%%%%%%%%%%%%%%%%%%%%%%%%%%%%
\makeatletter
\g@addto@macro\@maketitle{
\setcounter{figure}{0}
   \centering
    \includegraphics[width=\textwidth, trim=100 100 100 50, clip]{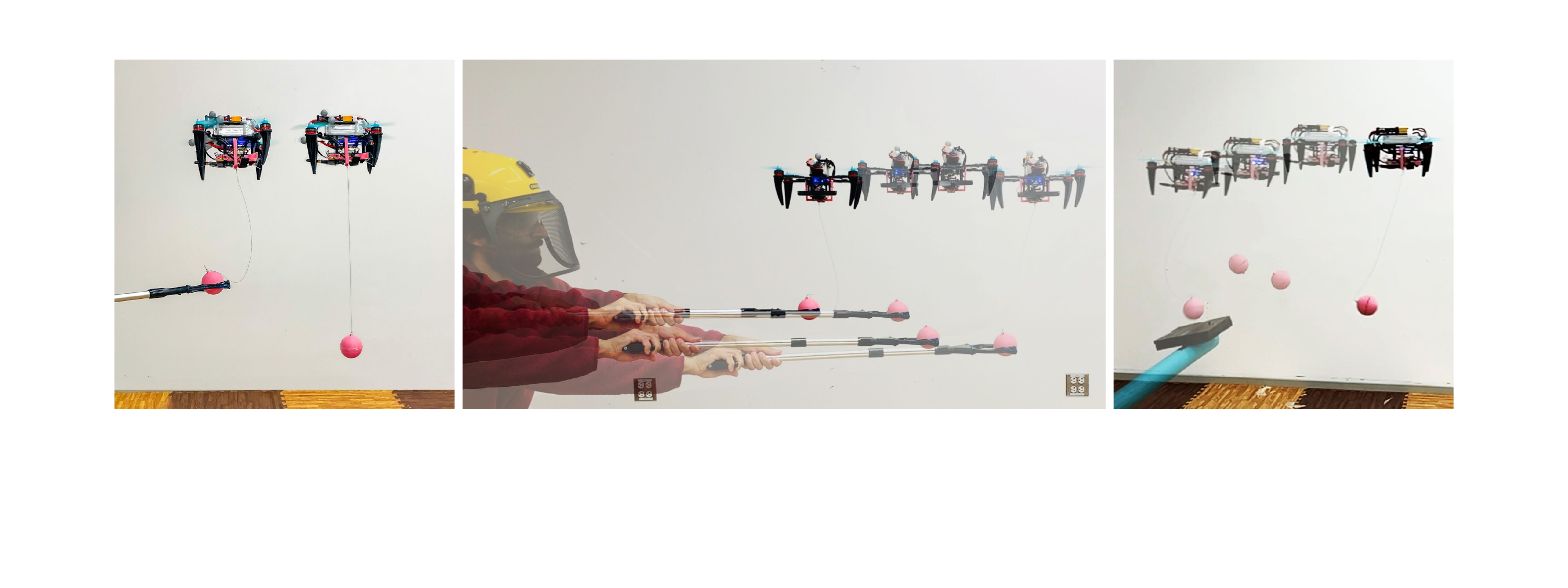} % Path to your image
    \vspace{-30pt}
    \captionof{figure}{Robot dual hybrid states (left). The robot on the right is in the taut state while the robot on the left is in the slack state. Snapshots showing a human-interaction task where the quadrotor tracks the load as a user manipulates it via a gripper (center). An experiment where we apply a vertical disturbance to a payload as the robot executes a trajectory (right).}

    \label{fig:motivation_pics}
}
\makeatother
\maketitle
\thispagestyle{empty}
\pagestyle{empty}

\begin{abstract}
Quadrotors equipped with cable-suspended loads represent a versatile, low-cost, and energy efficient solution for aerial transportation, construction, and manipulation tasks.
However, their real-world deployment is hindered by several challenges. The system is difficult to control because it is nonlinear, underactuated, involves hybrid dynamics due to slack-taut cable modes, and evolves on complex configuration spaces. Additionally, it is crucial to estimate the full state and the cable's mode transitions in real-time using on-board sensors and computation. To address these challenges, we present a novel Hybrid Perception-Aware Nonlinear Model Predictive Control (HPA-MPC) control approach for quadrotors with suspended loads. Our method considers the complete hybrid system dynamics and includes a perception-aware cost to ensure the payload remains visible in the robot's camera during navigation. Furthermore, the full state and hybrid dynamics' transitions are estimated using onboard sensors.
Experimental results demonstrate that our approach enables stable load tracking control, even during slack-taut transitions, and operates entirely onboard. The experiments also show that the perception-aware term effectively keeps the payload in the robot's camera field of view when a human operator interacts with the load. 

% To reliably control and deploy this system, it is important to design control and on-board estimation methods that account for the hybrid dynamics.
%Controlling a quadrotor with a cable-suspended payload is difficult due to its non-linear and under-actuated nature. The suspended payload mechanism increases the complexity of the dynamics by inducing dual motion modes. If the cable is taut, the payload exerts a force on the robot, and when the cable is slack, the robot and payload are decoupled. Controllers often do not account for hybrid mode transitions and instead ensure that the cable remains taut, reducing the agility and real-world practicality of the platform. To address this gap, we present a Hybrid Non-Linear Model Predictive Control (HNMPC) framework that includes state dependent costs to satisfy alternate objectives. Our experiments show that accounting for the hybrid mode enables stable control during slack-taut transitions of the system. We demonstrate that our controller can handle scenarios where external disturbances trigger unexpected mode transitions. We also show that perception awareness of the payload can be satisfied by including a cost term that prioritizes actions that keep the payload in the camera's field of view. Quadrotors are often compute-limited and this is an important limitation to consider when designing control algorithms. Unlike most works in this domain, we run the HNMPC and state estimation algorithms using onboard computation on a commercially available arm processor. 

\end{abstract}

\section*{Supplementary material}
\url{https://mrunaljsarvaiya.github.io/hpa-mpc.github.io/}

%%%%%%%%%%%%%%%%%%%%%%%%%%%%%%%%%%%%%%%%%%%%%%%%%%%%%%%%%%%%%%%%%%%%%%%%%%%%%%%%
\section{Introduction}
\IEEEPARstart{M}{icro} Aerial Vehicles (MAVs) play a pivotal role in a wide
range of applications such as surveillance and inspection \cite{trujillo2019}, search and rescue operations, and transportation~\cite{LoiannoRAL2018,sreenath2013geometric}. Additionally, they have recently become a popular choice for efficient and rapid transportation of supplies or materials compared to ground vehicles. For example, they are used during natural disasters \cite{BARMPOUNAKIS2016ijtst}, post-disaster relief operations, and the  equipment installation~\cite{Lindsey2012ConstructionWQ} in inaccessible areas. They are also energy efficient compared to other ground-based transportation solutions for last-mile delivery of small packages \cite{LastMileDelivery}. This makes them a valuable alternative transportation solution to reduce the environmental impact caused by millions of packages delivered every year.

Aerial robots can transport loads through active and passive manipulation mechanisms \cite{gowtham2015icra,guanrui2021ral}. Active mechanisms, such as grippers~\cite{Mellinger2013} and robot arms~\cite{afifi2023icuas}, enable precise manipulation but add hardware complexity and inertia therefore reducing agility, and increasing power consumption. In contrast, passive attachments, such as cables, magnets, and ball joints offer simplicity and reduced power requirements at the cost of increased control complexity. Cables are a particularly lightweight passive mechanism that provide a safe distance between the aerial robot and the payload, mitigating the payload's impact on the rotor's airflow and increasing safety when interacting with humans. We believe that the use of cables represents a favorable trade-off, balancing load maneuverability, manipulability, and safety while offering flexibility in executing multiple tasks~\cite{guanrui2024hri}.

However, developing control and estimation algorithms for a quadrotor with a cable-suspended payload is challenging due to its non-linear hybrid dynamics.
% evolving on either a $SE(3)\times S^2$ or $SE(3) \times \mathbb{R}^3$ manifold. 
Specifically, this system exhibits hybrid dynamical modes since the cable can be slack or taut as depicted in Fig.~\ref{fig:motivation_pics}. In the scenario shown in Fig.~\ref{fig:motivation_pics}, a human is interacting with the payload which can cause the cable to abruptly switch between the slack and taut modes. The state of the art model based control methods~\cite{sreenath2013geometric,guanrui2021pcmpc} that assume that the cable is always taut would fail in this scenario, as the cable tautness assumption is invalid. Additionally, if external disturbances such as wind or environment interactions cause the payload to move out of the onboard camera's Field of View (FoV), the state of the art estimation methods would also likely fail~\cite{sarah2018icra,guanrui2021pcmpc}. To address these challenges, we propose Hybrid Perception-Aware MPC (HPA-MPC) that actively promotes the visibility of the payload during navigation. We address the corresponding onboard state estimation problem to estimate the full hybrid dynamics therefore reducing the reliance on an external localization system.
%and actively keep the payload within the sensor's range. 

In this paper, we present the following contributions
\begin{itemize}
    \item We propose a novel Nonlinear Model Predictive Control (NMPC) method with dynamically updated cost functions to address the hybrid dynamics of a quadrotor with a cable-suspended payload. This solution is complemented by a novel state estimation approach based on an Extended Kalman Filter (EKF) technique, which detects the cable's hybrid mode and accurately estimates the payload's states, while relying solely on onboard sensors.
    \item We tackle the problem of payload visibility by introducing a perception-awareness cost function within the MPC framework. This promotes continuous tracking of the payload to keep it within the quadrotor's FoV, facilitating uninterrupted control once the cable becomes taut following an unexpected hybrid mode transition.
    \item We implement these methods on a quadrotor platform with Size, Weight, and Power (SWAP) limitations, achieving a navigation loop performance of $150$ Hz using purely onboard sensors and computing resources. We validate our proposed methods through extensive real-world experiments, including payload transportation, and human interactions tasks.

\end{itemize}

\section{Related Works} \label{sec:related_works}
Reliably controlling a slung payload system is challenging due to the dual hybrid modes of the system.  \cite{sreenath2013geometric,6631275} were the first works to discuss the hybrid dynamics and derive the system's differential flatness property. Initial research sidestepped the hybrid mode problem by tracking trajectories that explicitly constrain the cable to be taut or minimize payload swings \cite{FAUST2017381,6225213,GUERREROSANCHEZ2017433,7403277}. Limiting the payload swing is not only energy inefficient, requiring the quadrotor to counteract the forces exerted by the payload, but also restricts the ability to track aggressive and agile payload trajectories that undergo cable slack-taut transitions. 
% Most works either ignore the hybrid modes, rely on external motion trackers for state estimation or perform their computations offboard.

\cite{sarah2018icra} is one of the first papers to showcase aggressive flights with slung payloads and relies on results presented in \cite{sreenath2013geometric} 
and derives a load attitude controller using the system's differential flatness property. 
The computation is partially onboard as the approach utilizes an onboard vision payload detector but uses an external motion tracking for the quadrotor's state. The control system employs a payload attitude controller that unlike optimization-based controllers, does not account for state/input constraints. Therefore, the controller could demand infeasible outputs, such as a large change in angular velocities or thrust between subsequent controller cycles, which might be dynamically infeasible due to hardware constraints. 

\cite{guanrui2021pcmpc} employs Perception-Constrained MPC and similar to the proposed solution, uses onboard state estimation and control along with perception aware constraints. However, the approach does not account for the hybrid dynamics of the system. Moreover, introducing a perception constraint can lead to infeasible solutions as trajectory tracking may conflict with the perception objective. We circumvent this by formulating perception awareness as a weighted cost term.
Furthermore, \cite{falanga2018pampc} uses a perception aware cost to keep a stationary object in the robot's field of view during flight. However, in our case the object of interest is non-stationary and is a function of the robot's state.

The approach presented in \cite{haokun2024impact} is the most closely related to our Hybrid NMPC method. It accounts for the dual modes of the system and tracks trajectories that involve both dynamical modes. The trajectory generation algorithm produces trajectories that \textit{intentionally} switch modes whereas we focus on \textit{random} external disturbances that cause the cable to become slack. Furthermore, the proposed controller is only tested with an external motion capture system that provides near-perfect state and hybrid mode estimation. We demonstrate that our method can be reliably deployed on a system utilizing onboard vision-based state estimators equipped with payload perception awareness.
% Their work also relies on motion capture systems for robot and payload state estimation whereas our experiments only use onboard computation.

Most works that exploit the hybrid modes of this system focus on generating and tracking trajectories
\cite{7139492,8988166,foehn2017fast,unifiedControlQuad}. None of them validate their methods with onboard state estimators or discuss the controller's response if the system unexpectedly switches modes. Our experiments show that if a payload tracking controller does not account for hybrid modes, an unexpected mode switch would likely lead to a crash. Our HPA-MPC framework addresses this gap and increases the reliability of deploying quadrotors with slung payloads in environments prone to external disturbances.

\section{System Dynamics} \label{sec:system_dynamics}
\begin{figure}[t]
    \centering
    \includegraphics[width= \columnwidth, trim=290 420 660 170, clip]{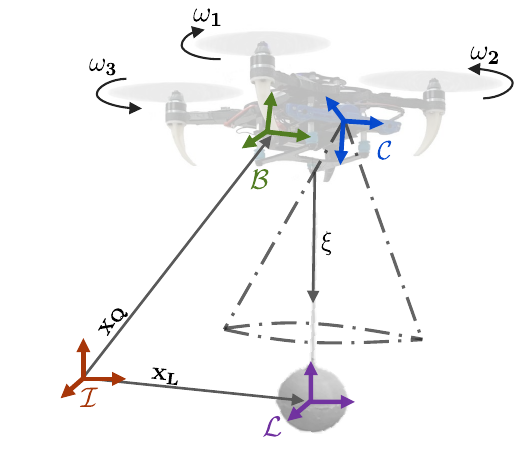} 
    \vspace{-5pt}
    \caption{System frame convention.\label{fig:frame_convention}}
\end{figure}

In this section, we present the hybrid dynamics model for a quadrotor with a suspended payload. A robot carrying a point mass payload is shown Fig.~\ref{fig:frame_convention} and the relevant variables are shown in Table \ref{tab:notation}. 

We first define the states and inputs of the system as 
\begin{align}
    \vecx &= \begin{bmatrix}\loadpos{}^\top, \loadvel{}^\top, \robotpos{}^\top, \robotvel{}^\top , \vecq{}^\top, \robotangvel^\top\end{bmatrix}^\top \label{eq:state_vector}, \\
    \vecu &= \begin{bmatrix}
    \omega_1, \omega_2, \omega_3, \omega_4
    \end{bmatrix}^\top,
\end{align}
where the states in $\vecx$ are the payload's and quadrotor's states defined in Table~\ref{tab:notation}, and $\omega_i$ are the motor speeds. The thrust generated by a single propeller $i$ is proportional to its squared motor speed as
\begin{equation}
    f_i = k_f {\omega}^{2}_i
\end{equation}

where $k_f$ is the motor constant. Therefore, the total thrust generated by 4 propellers is defined as 
\begin{align}
    f &= \sum_{i=1}^{4}k_f {\omega}^{2}_i,
\end{align}

The hybrid dynamics consists of two modes: i) taut mode, ii) slack mode. In each mode, the system has different dynamics and we summarize the corresponding equations of motion \cite{guanrui2024rotortm} below.
\subsubsection{Taut Mode} The system configuration space is $SE(3) \times S^2$. The cable imposes the constraint
\begin{equation} 
\loadpos = \robotpos{} + l\cablevec{} \label{eq:payload_quad_taut}
\end{equation}
where $\cablevec{} \in S^2$ is a unit vector from the robot's center of mass to the payload. By differentiating eq.~\eqref{eq:payload_quad_taut} we obtain
\begin{equation}
\cablevec = \frac{\loadpos - \robotpos{}}{l},~\cabledotvec=\frac{\loadvel -\robotvel{}}{l},~\cableddotvec{}=\frac{\loadacc{} - \robotacc{}}{l}    .\label{eq:cable_vel_acc_derivation}  
\end{equation}

\begin{table}[t]
\caption {Notation table\label{tab:notation}} 
\centering
%\newcolumntype{s}{}
\begin{tabularx}{0.48\textwidth}{>{\hsize=0.65\hsize}X >{\hsize=1.42\hsize}X}
    \hline \\
$\worldf$, $\loadf$, $\robotf{}$, $\cameraf$ & inertial, payload, robot  and camera frame\\
$\loadmass,m$ & mass of payload and robot\\
$\loadpos{},\robotpos{}$ & position of payload and robot in $\worldf$\\
$\loadvel,\loadacc$ & linear velocity, acceleration of payload in $\worldf$\\
$\robotvel,\robotacc$ & linear velocity, acceleration of robot in $\worldf$\\
$\robotrot{} \in\SOthree$& robot orientation with respect to $\worldf$ \\
$\robotangvel{}\in\realnum{3}$& angular velocity of robot in $\robotf{}$\\
$f_{}\in\realnum{}$, $\matM_{}\in\realnum{3}$&collective thrust and moment on robot in $\robotf{}$\\
$J \in \mathbb{R}^{3 \times 3}$ & moment of inertia of robot\\
$\cablevec{}\in S^2$&unit vector from robot to payload in $\worldf$\\
$l_{}, g\in\realnum{}$ & cable length, gravity constant\\
$\vecq_{}$& quaternion representation of $\robotrot{}$\\ \\
    \hline
\end{tabularx}
\end{table}

By following the derivation in \cite{sreenath2013geometric} and applying Lagrange-d'Alembert principle, we obtain

\begin{align}
\frac{d\loadpos}{dt} = \loadvel, ~\frac{d\robotpos{}}{dt} &= \robotvel{}, ~\dot{\vecq}_{} = \half\hat{\robotangvel{}}_{}\vecq_{},\label{eq:single-kinematics} \\
\prths{m+\loadmass}\prths{\loadacc + \vecg} &= \prths{\cablevec{}\cdot f\robotrot{}\axis{3}{}-m l\prths{\cabledotvec{}\cdot\cabledotvec{}}}\cablevec{},\label{eq:single-load-lagrange-eom} \\
m l\prths{\cableddotvec{}+\prths{\cabledotvec{}\cdot\cabledotvec{}}\cablevec{}} & = \cablevec{}\times\prths{\cablevec{}\times f\robotrot{}\axis{3}{}},\label{eq:single-quad-lagrange-eom}\\
\matM &= \inertia\robotangacc{} + \robotangvel{}\times\inertia\robotangvel{},\label{eq:single-robot-rotation-dyn}
\end{align}
where $\vecg = g \axis{3}{}$, $g = 9.81$ $m/s^{2}$ and $\axis{3}{} = \left[0~0~1\right]^{\top}$, and $\hat{\robotangvel{}}$ is the skew-symmetric matrix of the quadrotor angular velocity $\robotangvel{}$. We can write eqs.~\eqref{eq:payload_quad_taut}-\eqref{eq:single-robot-rotation-dyn} in the standard form as
\begin{equation}
\dot{\vecx}= g_p\prths{\vecx,\vecu},
\label{eq:taut_gen_form}
\end{equation}
where $g_p$ is the taut state dynamics function.

\subsubsection{Slack Mode} When the cable is slack, the payload's and the robot's motion are decoupled since the payload is in free fall and the rotors only effect the quadrotor's state. The system configuration space is now $SE(3) \times \mathbb{R}^3$ and the equations of motion are 
\begin{align}
\loadmass\prths{\loadacc + \vecg} &= 0,~f\robotrot{}\axis{3}{} = m\prths{\robotacc{} + \vecg},\label{eq:single-slack-dyn}\\
 \matM &=\inertia\angacc + \angvel\times\inertia\angvel \label{eq:single-slack-robot-rotation-dyn}.
\end{align}

We can now write eqs.~\eqref{eq:single-slack-dyn}-\eqref{eq:single-slack-robot-rotation-dyn} in the standard form as
\begin{equation}
\dot{\vecx} = g_z\prths{\vecx,\vecu},
\label{eq:slack_gen_form}
\end{equation}
where $g_z$ is the slack state dynamics function.

%  This plot has text slighly larger than system_frame_convention_4_smaller
% \begin{figure}[t]
%     \centering
%     \includegraphics[width= \columnwidth, trim=290 390 640 160, clip]{figures/system_frame_convention_4.pdf} 
%     \vspace{-5pt}
%     \caption{System frame convention.\label{fig:frame_convention}}
% \end{figure}

\section{Hybrid Perception Aware MPC (HPA-MPC)} \label{sec:hpampc}
\begin{figure*}
    \includegraphics[width=\textwidth, trim=200 500 370 400, clip]{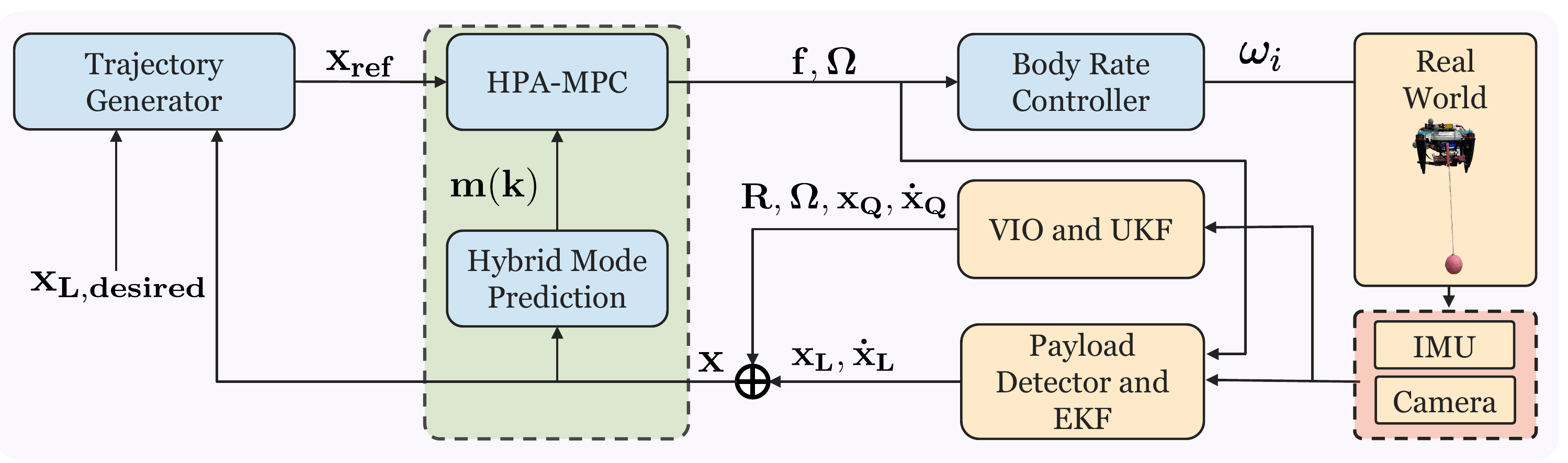}
    \caption{The system architecture describing the flow of information through the various components of our pipeline.
    \label{fig:sys_overview}}
\end{figure*}

We now present HPA-MPC which enables a quadrotor with a cable-suspended payload to track a desired payload trajectory, accounting for the system's complex hybrid dynamics. Our method actively keeps the payload within the robot's camera's FoV when external disturbances or interactions cause the cable to go slack, preventing the payload from moving out of the camera's range. The system overview of our solution is depicted in Fig. \ref{fig:sys_overview}.

\subsection{HPA-MPC Formulation} \label{subsec:hnmpc_formulation}
Our MPC solves the following control problem
\begin{subequations}
\begin{align}
\min_{\vecx(k), \vecu(k)}& \sum_{k=0}^{N-1} L(\vecx(k), \vecu(k), m(k)) \label{eq:nmpc_cost}\\
\text{subject to \, \, \,} &\vecx(k+1) = F(\vecx(k), \vecu(k), m(k)),\\
&\vecx(0) = \vecx_0, \\
&H_x(\vecx(k)) \leq 0, H_u(\vecu(k)) \leq 0\label{eq:nmpc_ineq_constraint_x}
\end{align}
\end{subequations}
where $N$ is the number of horizon steps, $k$ is the stage index and $H_x$ and $H_y$ are the set of inequality constraints on the states and inputs respectively. $m(k)$ is the slack-taut state of the system that is estimated via our onboard state estimation method. Specifically, $m(k) = 1$ implies that the system is taut and $m(k) = 0$ implies that the system is slack. The reader can refer to Section \ref{sec:state_estimation} for details of our approach to estimate the system state and detect hybrid mode transitions. $L$ is the stage cost and its implementation is discussed in Section \ref{subsec:cost_formulation}. $F$ is the discrete dynamics function that can be obtained using eqs in Section \ref{sec:system_dynamics} as 
\begin{equation}
F(\vecx(k), \vecu(k), m(k)) = 
\begin{cases} 
    G_p\prths{\vecx(k),\vecu(k)} & \text{if } m(k) = 1, \\
    G_z\prths{\vecx(k),\vecu(k)} & \text{if } m(k) = 0
\end{cases}
\end{equation}
where $G_p$ and $G_z$ are obtained by approximating eqs. \eqref{eq:taut_gen_form}, \eqref{eq:slack_gen_form} with an Implicit Runge–Kutta method. 

When the cable unexpectedly slackens due to an external force, the quadrotor loses control authority over the payload. If we assume the nature of the force (e.g., impulsive force), we can incorporate the force into the system dynamics and expect to account for it when solving the OCP. Instead, we make the \textit{alternate assumption} that the system will maintain its current hybrid mode throughout the MPC horizon. Since the MPC optimization frequency is $5\times$ the hybrid mode detector update rate, the controller can react to a hybrid mode switch with minimal delay.

% Therefore, if the system is slack, the payload is predicted to be in free fall for the entire horizon. To summarize
% \begin{equation}
%     m(k) = M, \text{ for } k \in (0, N),\label{eq:slack_taut_condition_1}
% \end{equation}
% where $M$ is the mode of the current system state.

% Optimizing the loss function should prioritize actions that ensure the predicted states track the reference trajectory. This is typically achieved by penalizing deviations of the predicted states from the reference values, which are generated by the trajectory generator through differential flatness. Therefore, we impose a cost on the state error $\tilde{\vecx}=\vecx_{ref,k} - \vecx_{k}$ and input error $\tilde{\vecu}=\vecu_{ref,k} - \vecu_{k}$, calculated by subtracting the predicted values from the reference trajectory $\vecx_{ref}$. As shown in Section \ref{subsec:hover_exp},  using the common payload tracking cost function when the cable is slack can cause the quadrotor to take actions leading to a crash. To mitigate this, we configure the cost function to ignore the payload tracking cost during a slack phase, allowing the quadrotor to track the nominal quadrotor trajectory, which is contained in $\vecx_{ref}$ and determined through differential flatness.
% We also demonstrate the feasibility of our system in achieving alternative objectives during the slack phase by incorporating a perception-aware cost to account for the camera's FOV. We conduct an experiment simulating a human-robot interaction scenario, where the payload is slack and continuously manipulated by an external agent.

\subsection{Cost Functions} \label{subsec:cost_formulation}
Our cost function is the sum of three terms
\begin{equation}
    L =  L_{\loadf} + L_{\robotf} + L_{\cameraf},  
\end{equation}  
where $L_{\loadf}$ represents the payload tracking cost, $L_{\robotf}$ the quadrotor tracking cost, and $L_{\cameraf}$ the perception-aware cost. $L_{\loadf}$, $L_{\robotf}$ are defined in terms of the state error $\tilde{\vecx}=\vecx_{ref,k} - \vecx_{k}$ and input error $\tilde{\vecu}=\vecu_{ref,k} - \vecu_{k}$, through the standard trajectory tracking cost formulation. $\vecx_{ref,k}$, $\vecu_{ref,k}$ denote the state and input reference and $\vecx_k$, $\vecu_k$ the current state and input.
% At every control cycle, the trajectory generator outputs the reference trajectory $\vecx_{ref}$ based on the desired payload trajectory $\vecx_{L}$. 
As illustrated in Section \ref{subsec:hover_exp}, utilizing a standard payload tracking cost function when the cable is slack may cause the controller to command actions that lead to a crash. To prevent this, we configure the cost function to ignore the payload tracking cost during a slack phase, enabling the quadrotor to track the nominal quadrotor trajectory.
We also demonstrate the system's capability to satisfy alternate objectives during the slack phase by integrating a perception-aware cost that considers the goal of keeping the load within the camera's FoV. We validate this approach by simulating a human-robot interaction task, where the payload is slack and continuously manipulated by an external agent.

\textit{1) Payload Tracking Cost:} This cost is active when the cable is taut $\Rightarrow m(k)=1$
\begin{align}
L_{\loadf} (k) =
m(k)( {\tilde{\vecx}(k)}^\top \mathbf{Q_{l}} {\tilde{\vecx}(k)} + {\tilde{\vecu}(k)}^\top \mathbf{R_{l}} {\tilde{\vecu}(k)}),
\end{align}
where $\mathbf{Q_{l}}$ and $\mathbf{R_{l}}$ are diagonal matrices. This term penalizes payload states, therefore the diagonal indices corresponding to quadrotor states are set to relatively small values. 

\textit{2) Quadrotor Tracking Cost:} This cost is active when the cable is slack $\Rightarrow m(k)=0$
\begin{equation} 
L_{\robotf} (k) =  
(1 - m(k))({\tilde{\vecx}(k)}^\top \mathbf{Q_{q}} {\tilde{\vecx}(k)}
+ {\tilde{\vecu}(k)}^\top \mathbf{R_{q}} {\tilde{\vecu}(k)}),
\end{equation}
where $\mathbf{Q_{q}}$ and $\mathbf{R_{q}}$ are diagonal matrices. This term penalizes quadrotor states, therefore the diagonal indices corresponding to payload states are set to zero. 

\textit{3) Perception-Aware Cost:} We would like this term to promote maximizing the camera's visibility of the payload, especially during a human-interaction task where the payload is more likely to go out of the camera's FoV. 
% One option is to penalize deviations of the projected payload coordinates from the image center.
% This can be achieved by normalizing  ${}^{\cameraf}x_{\loadf}$, ${}^{\cameraf}y_{\loadf}$ by ${}^{\cameraf}z_{\loadf}$   through the pin hole camera model. However, in our setup, the distance between the the camera and payload is bounded $\norm{\vecx_Q - \vecx_L} \leq l$ and hence the normalization can be avoided. 
We achieve this behavior by penalizing deviations of the $x$, $y$ payload position in the camera frame, ${}^{\cameraf}x_{\loadf}, {}^{\cameraf}y_{\loadf} $, from the camera origin.
% and ignore the ${}^{\cameraf}z_{\loadf}$ which create a less non-linear objective. 
This produces the cost function
\begin{align}
L_{cam} ={{}^{\cameraf}\mathbf{x}_{\loadf}}^\top \mathbf{Q_{cam}} {}^{\cameraf}\mathbf{x}_{\loadf},\label{eq:pampc_cost}
\end{align}

All the optimization variables need to be expressed in terms of the state vector defined in eq.~\eqref{eq:state_vector}. Exploiting several relations, we obtain the following result 
\begin{align}
{}^{\cameraf}\mathbf{x}_{\loadf} &= {}^{\cameraf}  \mathbf{T}_{\worldf} \mathbf{x}_{\loadf} = {{}^{\worldf} \mathbf{T}_{\cameraf}}^{-1} \mathbf{x}_{\loadf} \\
% &= {({}^{\worldf} \mathbf{R}_{\cameraf})}^{-1} 
%     (\loadpos{} - \mathbf{R} {}^{\robotf} \mathbf{t}_{\cameraf} + \robotpos{}) \\
&= {(\mathbf{R} {}^{\robotf} \mathbf{R}_{\cameraf})}^{-1} 
    (\loadpos{} - \mathbf{R} {}^{\robotf} \mathbf{t}_{\cameraf} + \robotpos{}),
\end{align}
which expresses ${}^{\cameraf}\mathbf{x}_{\loadf}$ in terms of the state variables $\loadpos{}$, $\robotpos{}$, $\mathbf{R}$ and the fixed rotation ${}^{\robotf} \mathbf{R}_{\cameraf}$ and translation ${}^{\robotf} \mathbf{t}_{\cameraf}$ between the robot and camera frame. $\mathbf{{}^{\mathcal{A}}T_{\mathcal{B}}}$ denotes the transformation matrix of frame $\mathcal{B}$ with respect to frame $\mathcal{A}$. 

\subsection{Advantages of HPA-MPC} \label{subsec:benefits_over_alt}
Our method provides an optimization framework capable of predicting and exploiting the hybrid modes during the optimization process and supports custom state dependent cost functions. An alternate method to handle the hybrids modes is to use a non-hybrid payload NMPC in the taut state and switch to a geometric controller in the slack state. We argue that the benefits of integrating the hybrid states into a single NMPC model include
\begin{enumerate}
    \item This framework more effectively handles impact forces generated during transitions from slack to taut states. By modelling the disturbance and including it in the dynamics model, the NMPC could directly predict the slack-taut transition using  eq.~\eqref{eq:slack_taut_condition_2}.
    % , since the state vector includes $\loadpos$ and $\robotpos$.
    \item Using the NMPC framework in the slack state allows for the addition of custom cost functions to achieve alternative goals. We demonstrate this by incorporating perception awareness during a payload handling task.
\end{enumerate}

\section{State Estimation} \label{sec:state_estimation}
% In this section, we present the state estimation method, the payload detector and hybrid state estimator using onboard sensors. 
To estimate the state of the payload, we estimate the quadrotor state in the $\worldf$ frame using VIO and leverage that information to estimate the $\loadpos$ and $\loadvel$ by extracting features from the camera pointing at the payload.

\subsubsection{Visual Inertial Odometry}
The robot's pose is estimated using a downward facing monocular camera and an IMU. We use an Unscented Kalman Filter (UKF) to combine the VIO measurements with the acceleration and gyro data from the IMU which gives state prediction updates at our IMU frequency of $500$ Hz. We point the reader to \cite{loianno2017ral, guanrui2021pcmpc} for additional details on this approach. 

\subsubsection{Payload State Estimation} \label{subsec:state_estimation_payload}
% \begin{figure}[htbp]
%     \centering
%     \begin{minipage}[b]{0.48\columnwidth}
%         \centering
%         \includegraphics[width=\textwidth, trim=200 0 200 100, clip]{figures/load_det_1.png} % Path to your image
%     \end{minipage}
%     \begin{minipage}[b]{0.48\columnwidth}
%         \centering
%         \includegraphics[width=\textwidth, trim=200 0 200 60, clip]{figures/load_det_2.png} % Path to your image
%     \end{minipage}
%     \caption{Raw camera image (left) and the detected payload boundary (right)}
%     \label{fig:payload_cam}
% \end{figure}

To estimate the payload position and velocity, we extract camera measurements, $\loadpos$, at $30$ Hz. To do this, we detect the boundary of the spherical payload in the camera image using OpenCV and then use the ellipse fitting algorithm defined in~\cite{justin2017ral} to estimate ${}^\cameraf \loadpos$. 
Finally, we find $\cablevec{}$ and $\cabledotvec{}$ using eqs.~\eqref{eq:cable_vel_acc_derivation}, \eqref{eq:cam_paylaod_tf_2}
\begin{equation}
\loadpos = {}^{\worldf}  \mathbf{T}_{\cameraf} {}^\cameraf \loadpos = {}^{\worldf} \mathbf{T}_{\robotf}{}^{\robotf}  \mathbf{T}_{\cameraf} {}^\cameraf \loadpos \label{eq:cam_paylaod_tf_2}.
\end{equation}
To obtain higher frequency updates, we plug $\cablevec{}$, $\cabledotvec{}$ into an Extended Kalman filter (EKF) that estimates the cable direction and velocity using the camera measurements, IMU and commanded motor thrust. Once we have $\cablevec{}$, we can find $\loadpos$ using eq. \eqref{eq:payload_quad_taut}. We summarize the EKF equations presented in \cite{guanrui2021pcmpc} below. The filter state and inputs are 
\begin{equation}
    \matX =  \begin{bmatrix} \cablevec{} & \cabledotvec{} \end{bmatrix},~ 
    \matU = f.
\end{equation}
The process model is defined as
\begin{equation}
\dot{\mathbf{X}} = \begin{bmatrix}
    \cabledotvec{} \\
    \frac{1}{m_q l} \cablevec{} \times (\cablevec{} \times \mathbf{u}) - \left\lVert \cabledotvec{} \right\rVert_2^2 \cablevec{}
\end{bmatrix} + \mathbf{N},
\end{equation}
The measurement model is defined as 
\begin{equation}
\mathbf{Z} = \begin{bmatrix}
    \mathbf{p}^B \\
    \dot{\mathbf{p}}^B
\end{bmatrix} = g(\mathbf{X}, \mathbf{V}) = \begin{bmatrix}
    \mathbf{R}^\top l \cablevec{} \\
    l \left( \mathbf{R}^\top \cabledotvec{} - \robotangvel{} \times \mathbf{R}^\top \cablevec{} \right)
\end{bmatrix} + \mathbf{V},
\end{equation}
The process noise \(\mathbf{N} \in \mathbb{R}^6\) is as additive Gaussian white noise \(\mathbf{N} \sim \mathcal{N}(0, \mathbf{Q_N})\) with zero mean with covariance \(\mathbf{Q_N} \in \mathbb{R}^{6 \times 6}\). Similarly, the measurement noise is additive Gaussian white noise 
\(\mathbf{V} \sim \mathcal{N}(0, \mathbf{Q_V})\) with zero mean and standard deviation \(\mathbf{Q_V} \in \mathbb{R}^6\).

% Note that an additional benefit of using a Kalman Filter for payload state estimation is that if the payload unexpectedly moves out of the FOV, the filter will continue publishing new estimates using the sensor readings. 

\subsubsection{Hybrid Mode Detector} \label{sec:mode_detector}
It is important to identify the hybrid modes of the system with low latency so HPA-MPC can use the appropriate hybrid dynamics, especially while executing a trajectory. Once we have $\vecx_Q$ and $\vecx_L$ using the methods above, we estimate the hybrid mode using
\begin{equation}
m(k) =
\begin{cases} 
0 & \text{if } \left\| \loadpos(k) - \robotpos{}(k) \right\| < l - \epsilon\\ \label{eq:slack_taut_condition_2}
1 & \text{else } 
\end{cases}
\end{equation}
where $\epsilon=0.05~\si{m}$ and was experimentally derived to mitigate the effects of measurement inaccuracies that could cause false slack mode predictions. Additionally, we use a low pass filter to reduce high frequency noise in the estimated cable length.

% Note that when the payload is slack, the filter doesn't produce accurate updates. To address this, the filter uses the last measurement update to determine if the cable is slack. Specifically, if $\left\| \loadpos - \robotpos{} \right\| < l$, it directly publishes the measurement updates without applying the process model. We acknowledge this limitation in our state estimation system and aim to address it in future work by enabling the filter to use the appropriate dynamics model in slack scenarios.
% It is important to note that this assumption does not hold in all scenarios, such as immediately before the cable transitions from a slack to a taut state. However, our experiments demonstrate that, despite this discrepancy, our controller manages slack-to-taut transitions more effectively than a non-hybrid controller.

\section{Experimental Results}  \label{sec:experiments}
\begin{figure*}[t]
    \begin{minipage}[b]{\textwidth}
        \centering
        \vspace{-5pt}
        % \hspace{10pt}
        \includegraphics[width=\textwidth, trim=20 30 20 0, clip]{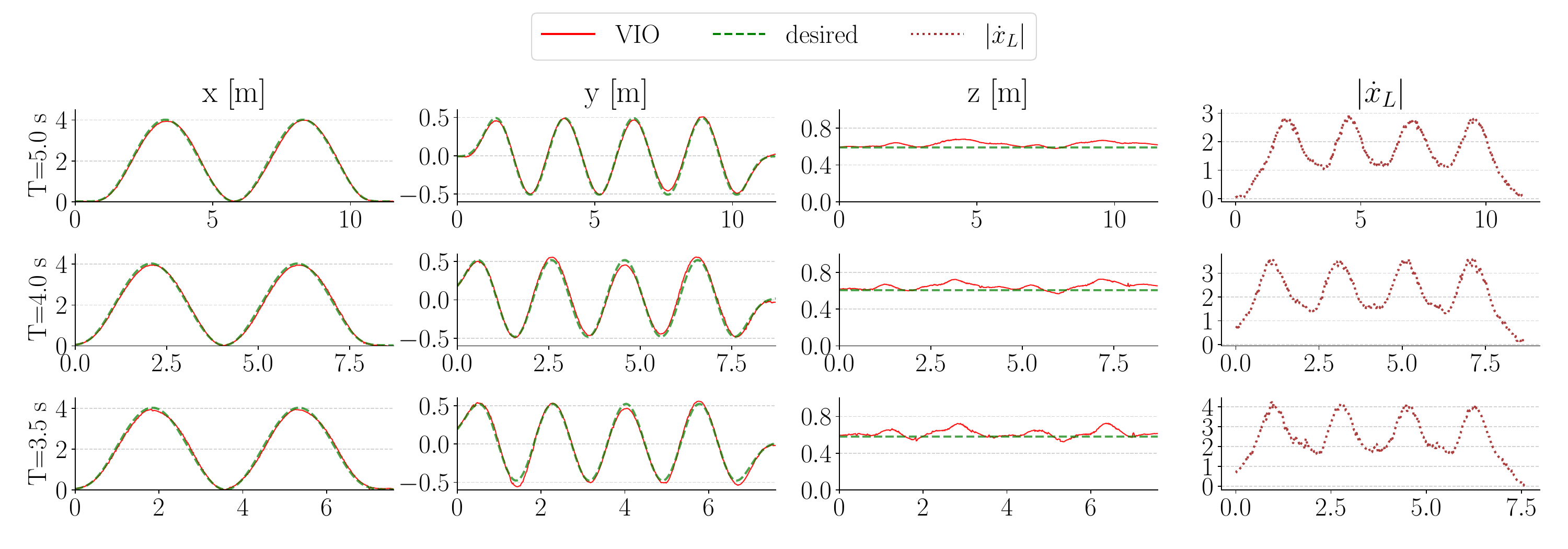}
        \label{fig:tracking_stats}
    \end{minipage}
    
    \begin{minipage}[b]{0.65\columnwidth}
        \centering
        \includegraphics[width=\columnwidth,  trim=50 0 120 50, clip]{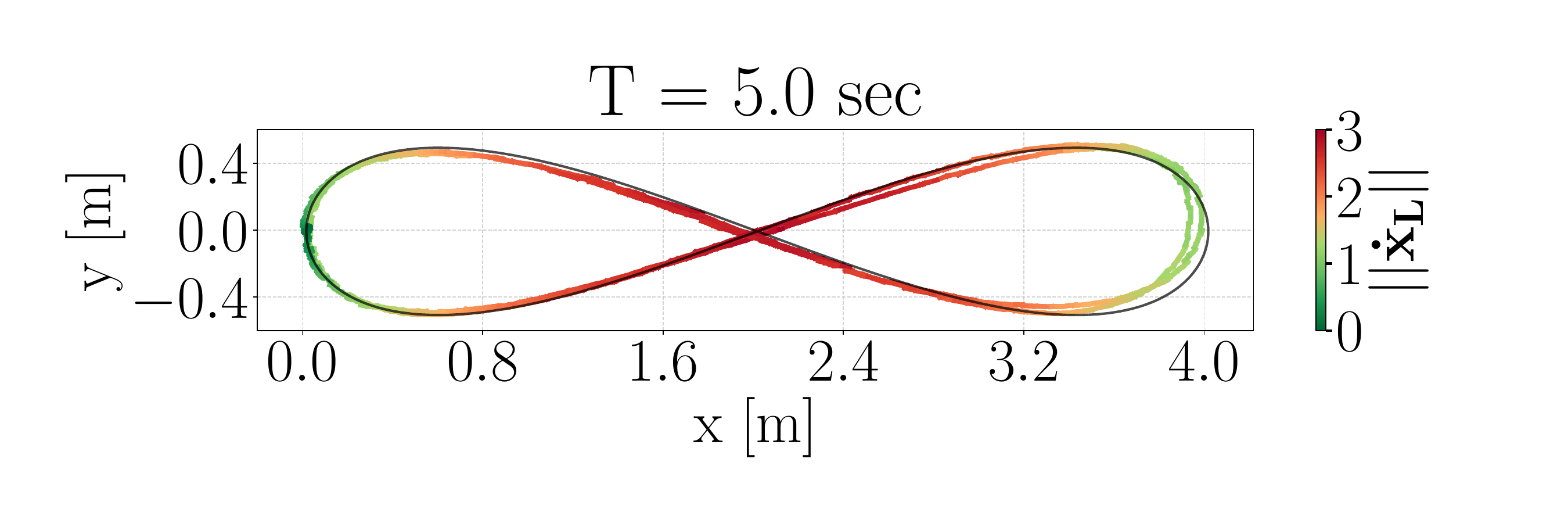}
    \end{minipage}
    \hfill
    \begin{minipage}[b]{0.65\columnwidth}
        \centering
        \includegraphics[width=\columnwidth,  trim=50 0 120 50, clip]{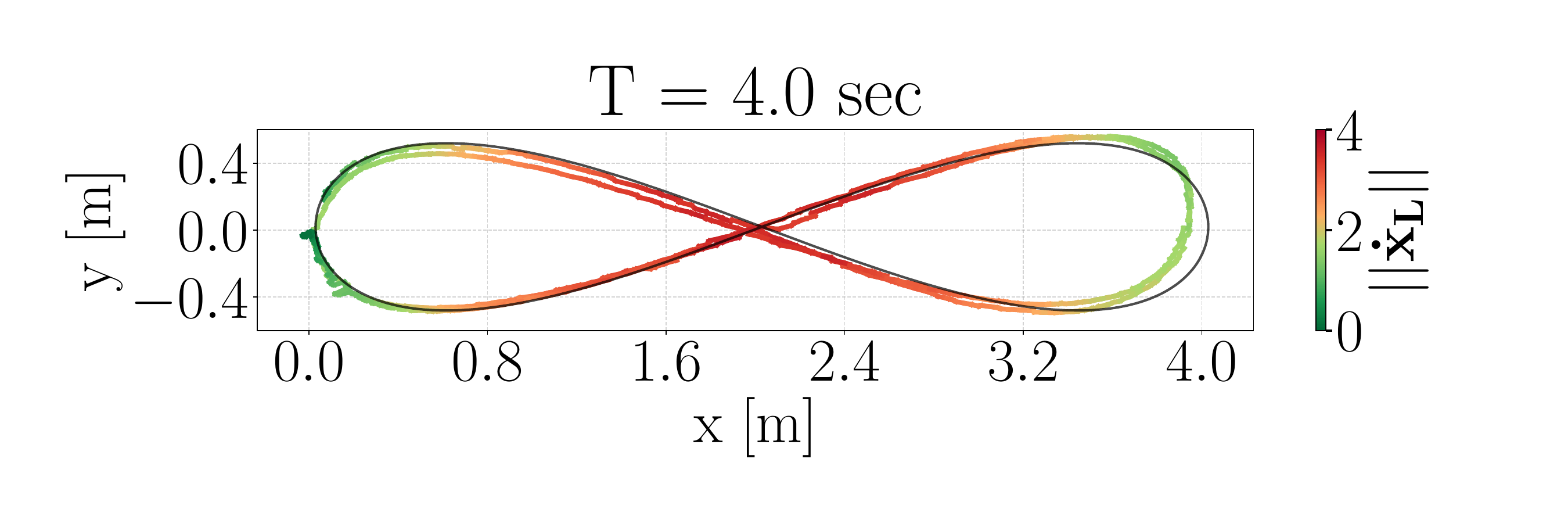}
    \end{minipage}
    \hfill
    \begin{minipage}[b]{0.65\columnwidth}
        \centering
        \includegraphics[width=\columnwidth,  trim=50 0 120 50, clip]{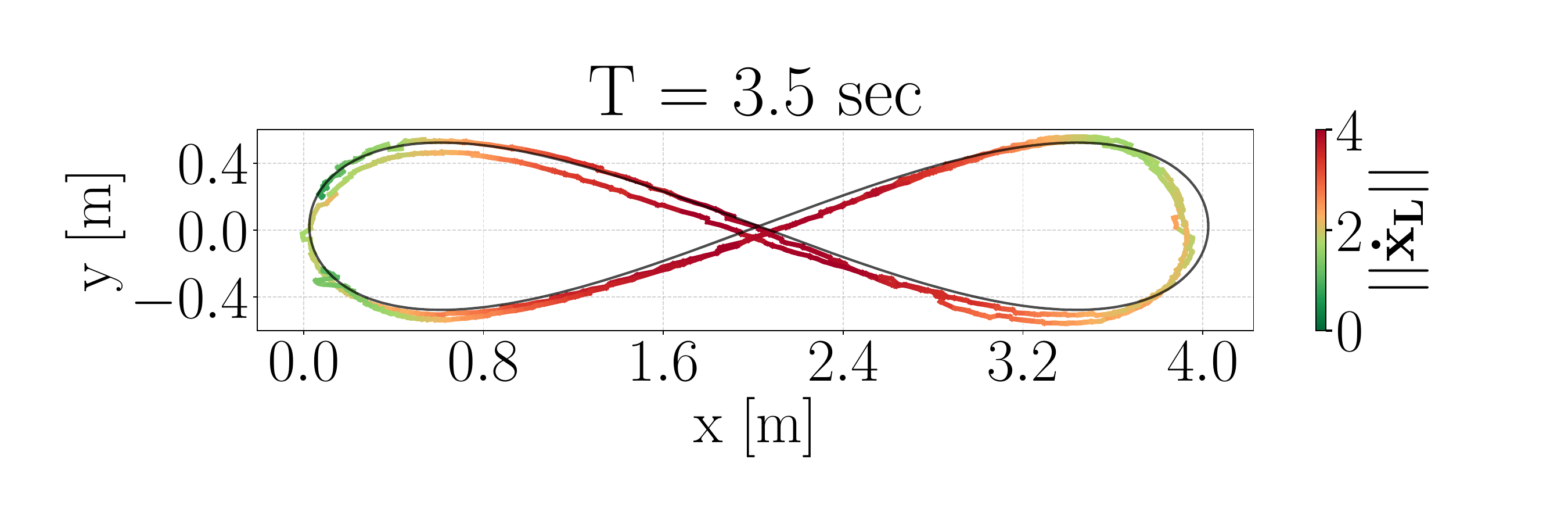}
    \end{minipage}
    \hfill
    \caption{Payload trajectory tracking plots for following a lissajous trajectory. The last row shows the $x-y$ visualization of payload path with a colorbar indicating the magnitude of the payload linear velocity.}  
    \label{fig:liss_tracking_res}
    %\vspace{-10pt}
\end{figure*}

We present the experimental validation of our proposed HPA-MPC and state estimation methods. We first demonstrate the effectiveness of our approach by showing trajectory tracking results. Next, we show our method's ability to detect the slack and taut states of the cable and handle the instantaneous impacts on the quadrotor generated when the cable transition from slack to taut. Furthermore, to show the importance of accounting for hybrid dynamics, 
% in control and estimation, 
we compare our method with a MPC and estimation methods that assume the cable is always taut during hovering and trajectory tracking tasks. External disturbances are introduced to the payload to induce slack-taut transitions, and the experimental results show that neglecting the hybrid state leads to crashes, proving the necessity of our approach. Lastly, we validate the perception-aware cost function in HPA-MPC, demonstrating the quadrotor's ability to actively track the payload by keeping it within its FoV during human interaction tasks where the cable is intentionally slackened.

The experiments are conducted in an indoor testbed
% with a flying space of $10\times6\times4~\si{m^3}$ 
of the Agile Robotics and Perception Lab (ARPL) at New York University. The quadrotor uses a $\text{Modalai}^{\circledR}\text{VOXL2}^{\text{TM}}$ board and is equipped with a Qualcomm QRB5165 processor and a PX4 Autopilot. The processor uses an arm Kryo 585 CPU with three core types: 4 low-speed cores, 3 medium-speed cores, and 1 high-speed core. The robot is equipped with an IMU and two monocular cameras. A $1280 \times 720$ pixels camera to estimate the payload state, and a $640 \times 480$ pixels camera for VIO. The quadrotor weighs $0.72~\si{kg}$, the payload weighs $0.1~\si{kg}$ and the cable length is $0.5~\si{m}$. 

\subsection{Software Setup}
Our software is implemented in ROS2 using C++. We run our HPA-MPC algorithm onboard at $150$ Hz and solve the OCP using acados \cite{verschueren2022acados}. We set the time horizon to $1$ s and $N=10$ steps. We initialize the first iteration of the solver with the ideal hover thrust command and zero angular velocities and use the previous cycle's solution to warm start the next iteration. To achieve a high control frequency, we utilize acados's RTI scheme that runs a single SQP cycle every iteration. This trades off the quality of the solution for speed \cite{verschueren2022acados}. The predicted angular velocities $\robotangvel{}$ and collective thrust $f$ of the MPC's second horizon stage are sent to the low level PX4 angular velocity controller. We are able to run MPC at $150$ Hz on this computationally constrained platform by configuring acados to use OpenMP \cite{dagum1998openmp} and restricting it to use the medium and high-speed cores. 

\subsection{Payload Trajectory Tracking}
We validate our HPA-MPC method to track Lissajous payload trajectories that are generated using the equation

\begin{equation}
\loadposdes(t) = \begin{bmatrix}
    a\sin(t)&b\sin(nt + \phi)&c\sin(mt) + \psi    \label{eq:lissajous_traj_eq}
\end{bmatrix}^\top,
\end{equation}
where $n,m$ control the relative periods of $x,y,z$ motions and $a,b$ and $c$ their relative amplitudes. By differentiating eq.~\eqref{eq:lissajous_traj_eq}, we obtain $\loadveldes(t)$ and $\loadaccdes(t)$. Fig. \ref{fig:liss_tracking_res} and Table \ref{tab:tracking_error} show the results for tracking a lissajous trajectory with $a=2$, $b=0.5$, $n=2$.

\begin{table}[htbp]
%\vspace{-10pt}
\caption {Payload trajectory tracking Root Mean Square Error (RMSE).\label{tab:tracking_error}} 
\centering
%\newcolumntype{s}{}
\begin{tabularx}{\columnwidth}{>{\hsize=0.375\hsize}X>{\hsize=0.1\hsize}X>{\hsize=0.25\hsize}X>{\hsize=0.25\hsize}X>{\hsize=0.25\hsize}X}\hline
 \rule{0pt}{2ex} & &T = 5s  & T = 4s & T = 3.5s\\\hline
 Position [m]& $x$    & 0.052    &  0.068 &  0.073 \\
        & $y$    & 0.021    &  0.032  &  0.027\\
        & $z$    & 0.039    &  0.044  &  0.043\\\hline
Max $|\dot{\vecx}_L|$ [m/s]& \text{}    & 3.0    &  3.6 &  4.2 \\
    \hline
\end{tabularx}
\vspace{-10pt}
\end{table}

\subsection{Slack-Taut Transition} 
\subsubsection{Hover}\label{subsec:hover_exp}
\begin{figure*}[t]
    \centering
    \begin{minipage}[b]{0.14\columnwidth}
         \begin{flushleft}
            \includegraphics[width=\textwidth, trim=335 0 335 0, clip]{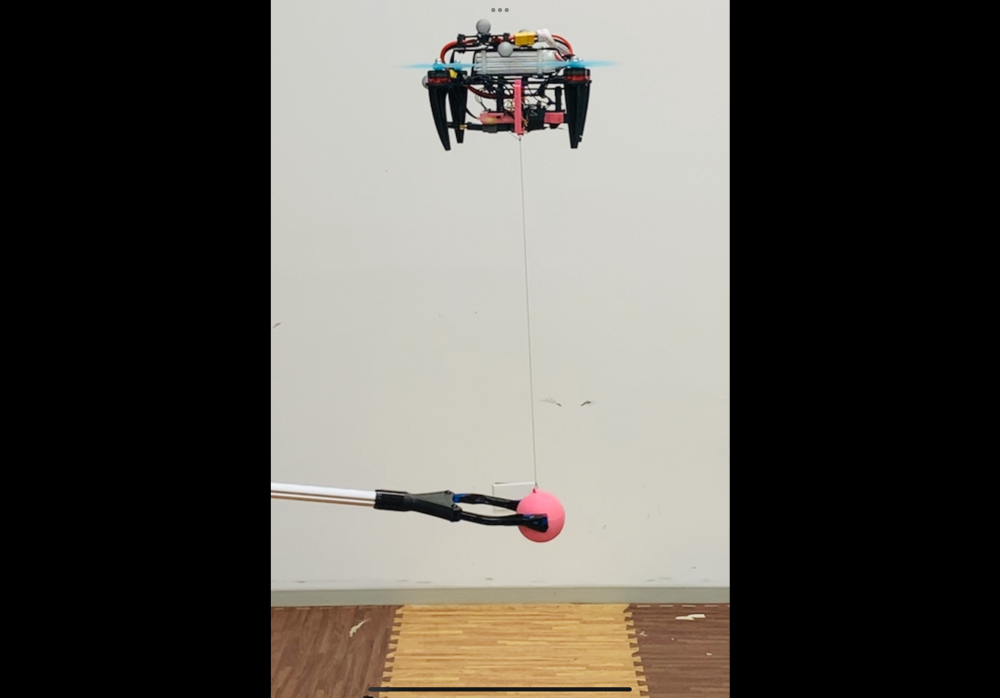} % image
        \end{flushleft}
    \end{minipage}
    \hfill
    \begin{minipage}[b]{0.14\columnwidth}
        \centering
        \includegraphics[width=\textwidth, trim=335 0 335 0, clip]{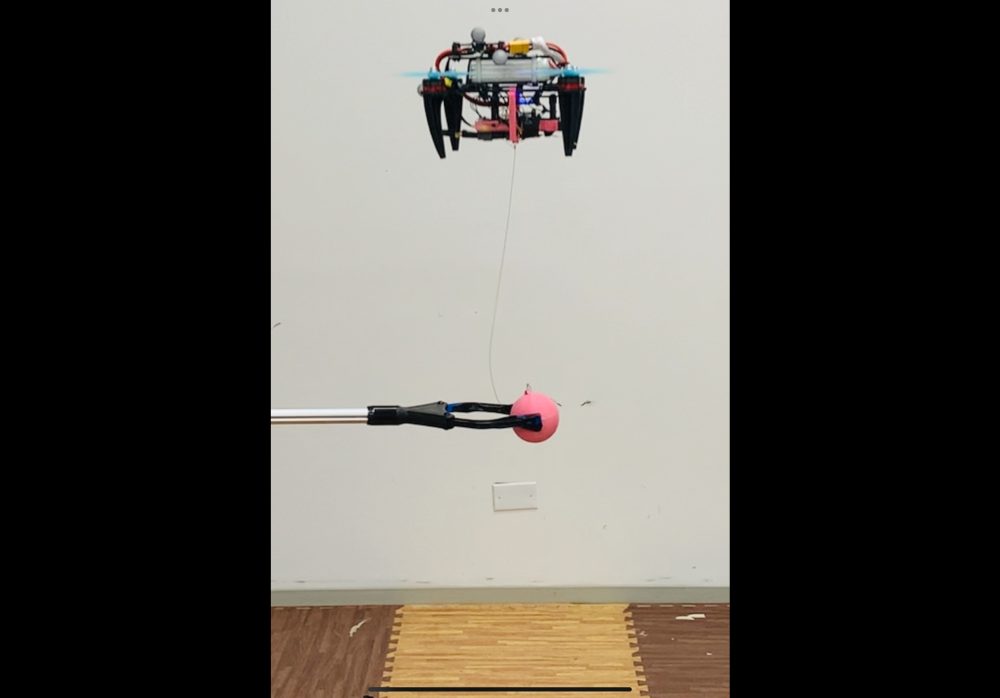} % image
    \end{minipage}
    \hfill
    \begin{minipage}[b]{0.14\columnwidth}
        \centering
        \includegraphics[width=\textwidth, trim=335 0 335 0, clip]{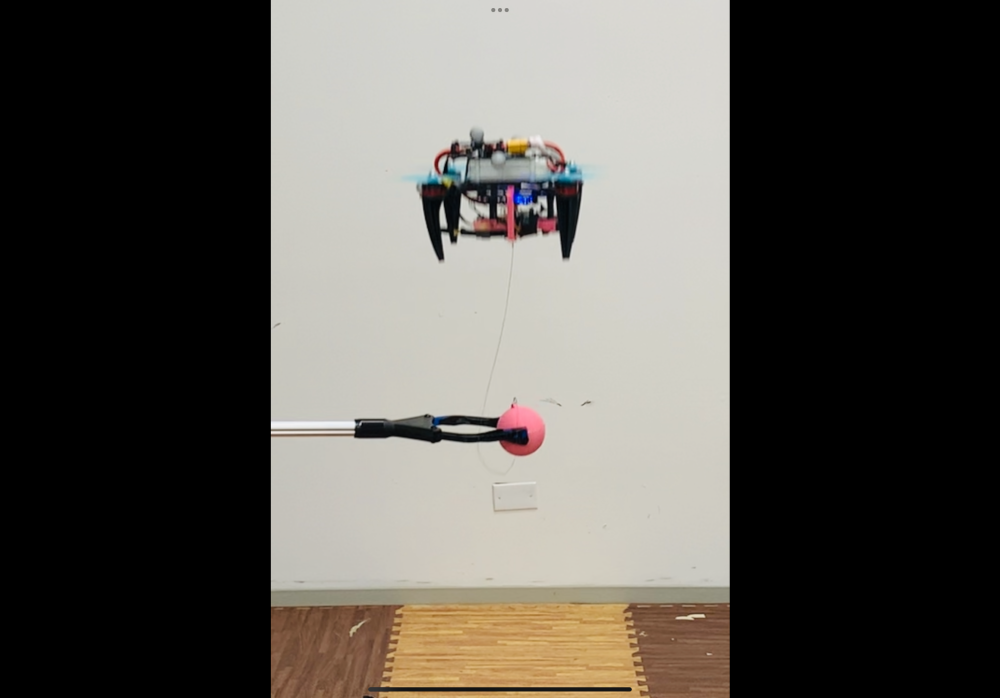} % image
    \end{minipage}
    \hfill
    \begin{minipage}[b]{0.14\columnwidth}
        \centering
        \includegraphics[width=\textwidth, trim=335 0 335 0, clip]{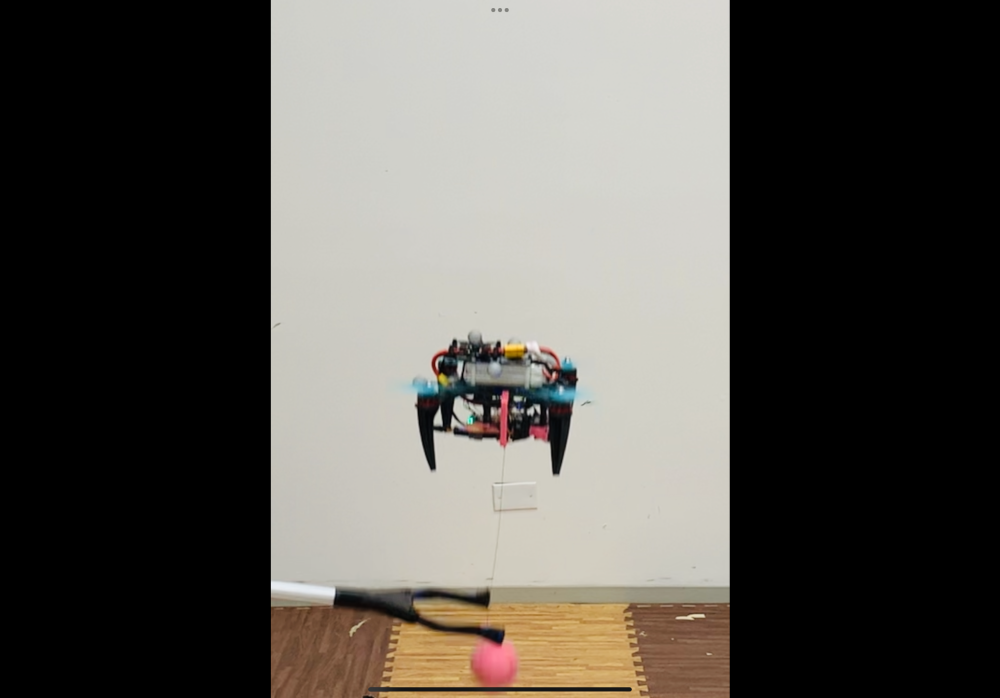} % image
    \end{minipage}
    \hfill
    \begin{minipage}[b]{0.14\columnwidth}
        \centering
        \includegraphics[width=\textwidth, trim=335 0 335 0, clip]{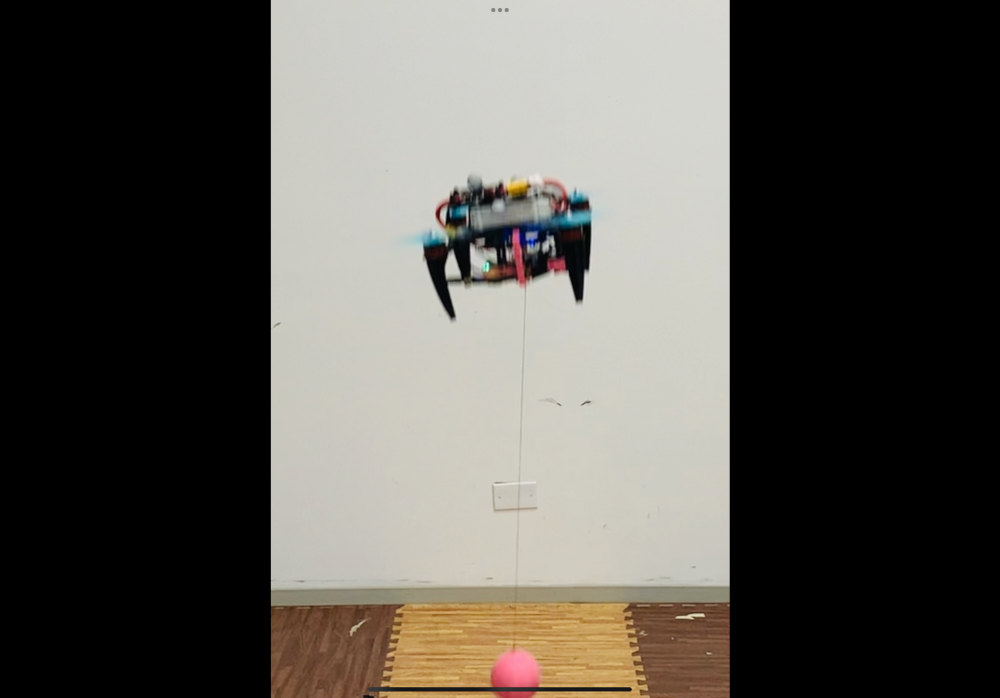} % image
    \end{minipage}
    \hfill
    \begin{minipage}[b]{0.14\columnwidth}
        \includegraphics[width=\textwidth, trim=335 0 335 0, clip]{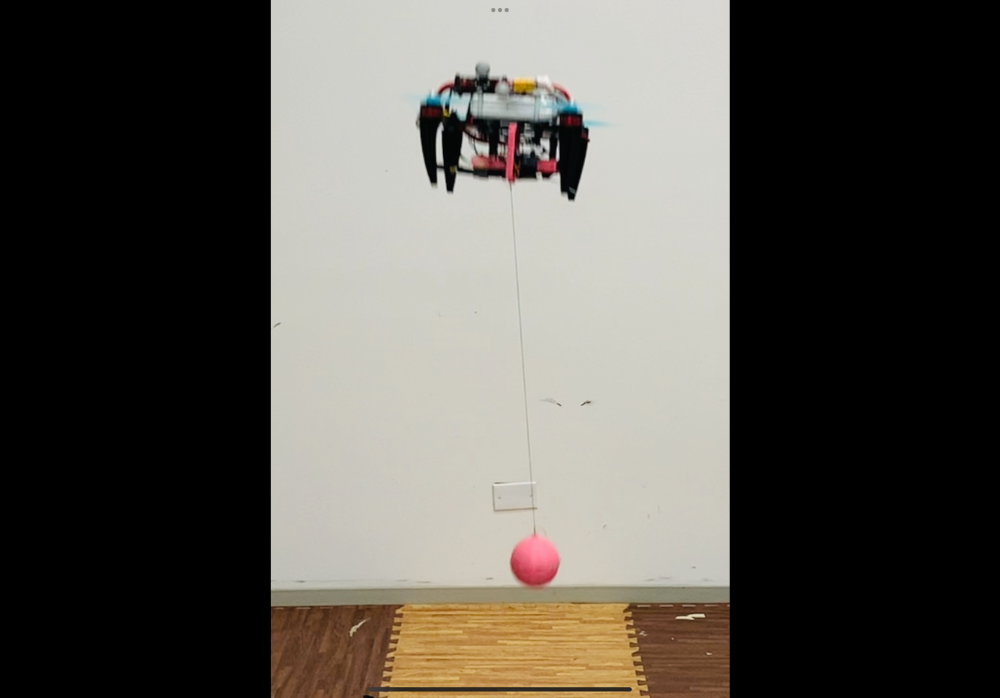} % image
    \end{minipage}
    %  seconf gif
    \hspace{20pt}
    \begin{minipage}[b]{0.14\columnwidth}
        \centering
        \includegraphics[width=\textwidth, trim=100 40 100 40, clip]{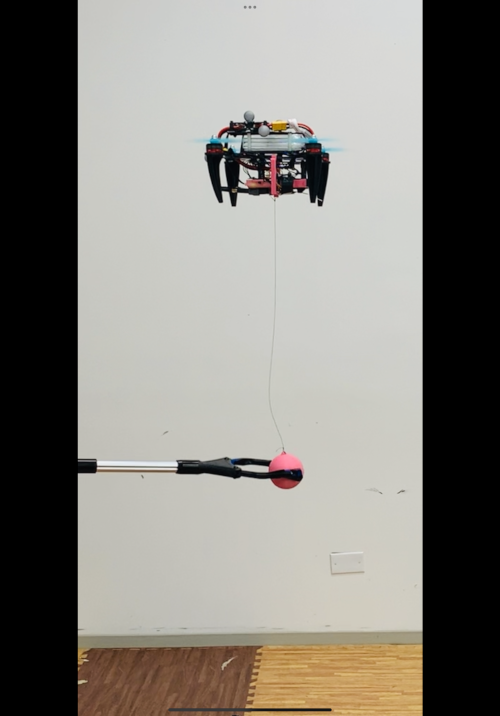} % image
    \end{minipage}
    \hfill
    \begin{minipage}[b]{0.14\columnwidth}
        \centering
        \includegraphics[width=\textwidth, trim=100 40 100 40, clip]{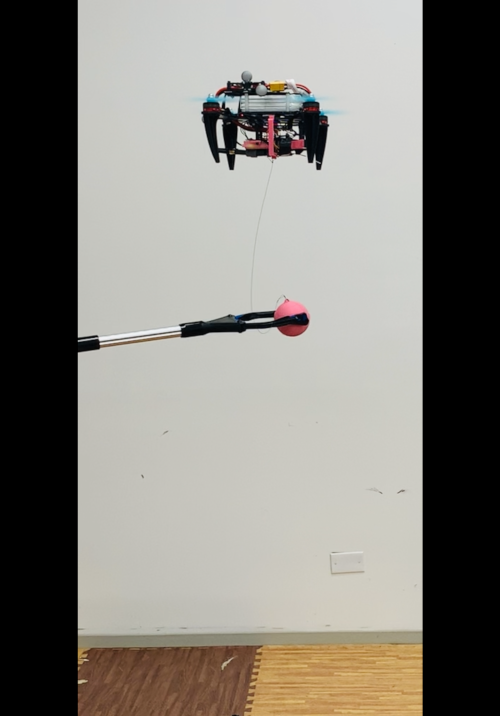} % image
    \end{minipage}
    \hfill
    \begin{minipage}[b]{0.14\columnwidth}
        \centering
        \includegraphics[width=\textwidth, trim=100 40 100 40, clip]{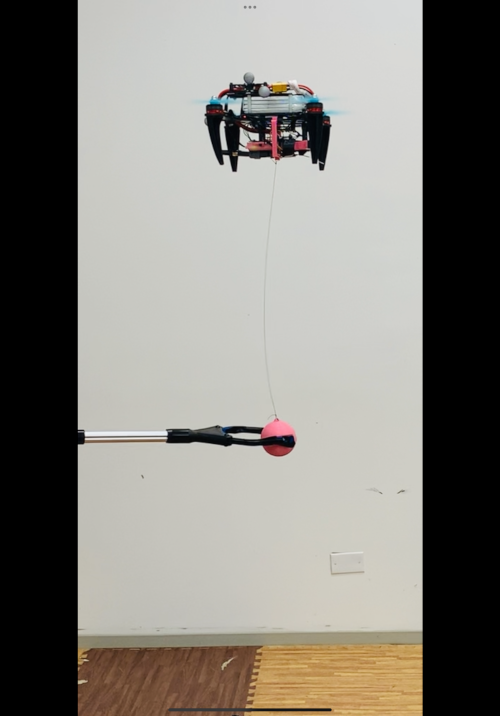} % image
    \end{minipage}
    \hfill
    \begin{minipage}[b]{0.14\columnwidth}
        \centering
        \includegraphics[width=\textwidth, trim=100 40 100 40, clip]{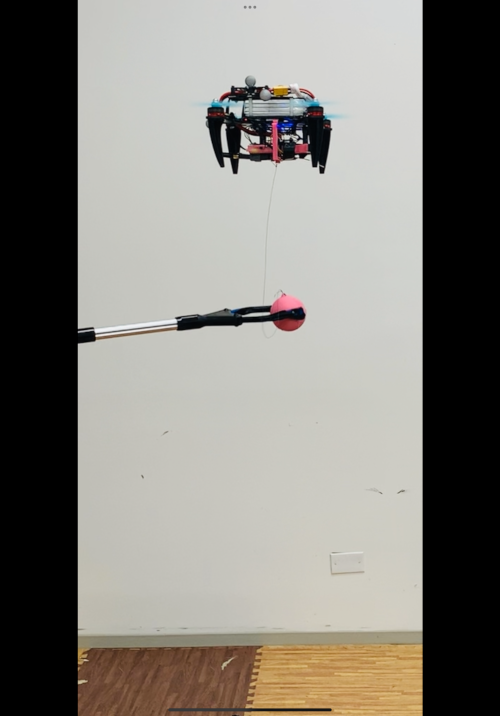} % image
    \end{minipage}
    \hfill
    \begin{minipage}[b]{0.14\columnwidth}
        \centering
        \includegraphics[width=\textwidth, trim=100 40 100 40, clip]{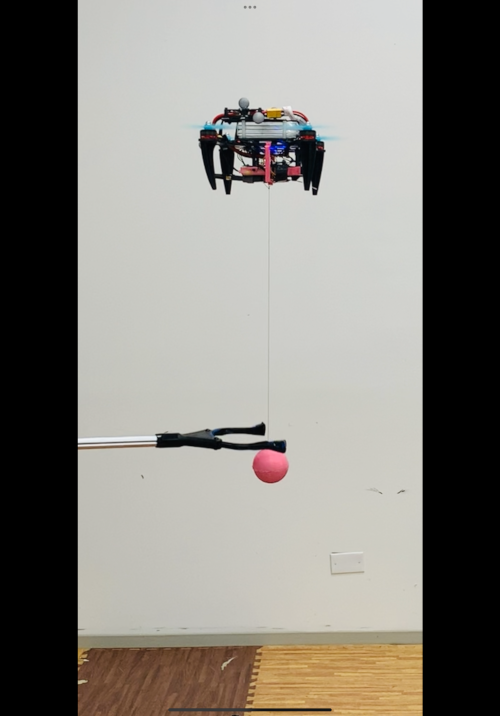} % image
    \end{minipage}
    \hfill
    \begin{minipage}[b]{0.14\columnwidth}
        \centering
        \includegraphics[width=\textwidth, trim=100 40 100 40, clip]{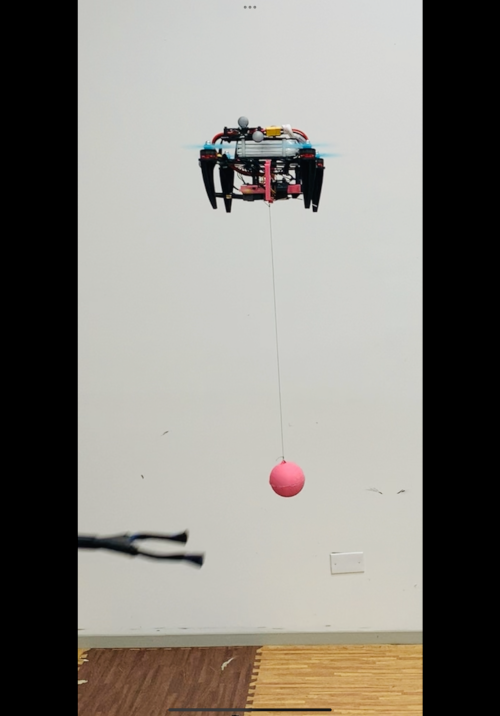} % image
    \end{minipage}
    \vspace{10pt}
    \begin{minipage}[b]{\columnwidth}
        \hspace{5pt}
        \includegraphics[width=\columnwidth, trim=20 0 50 0, clip]{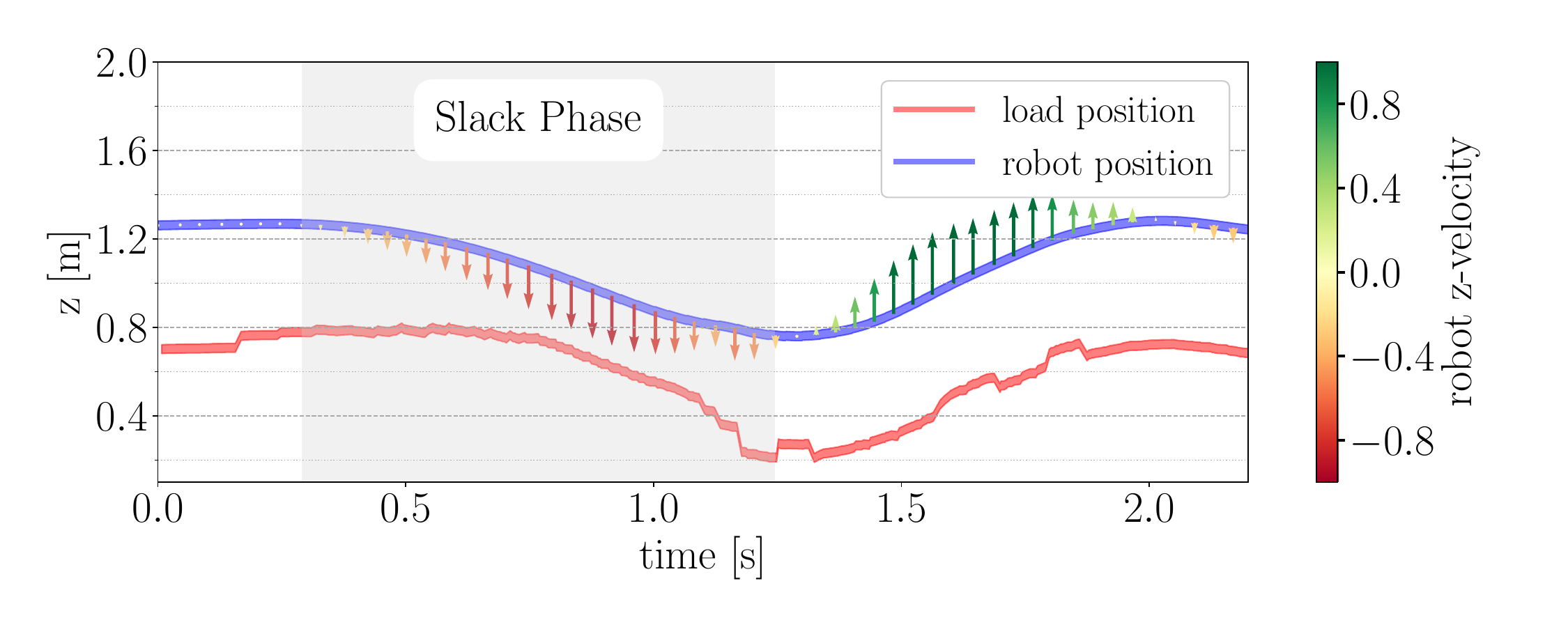}

    \end{minipage}
    \hspace{5pt}
    \begin{minipage}[b]{\columnwidth}
        \centering
        \includegraphics[width=\columnwidth, trim=20 0 80 0, clip]{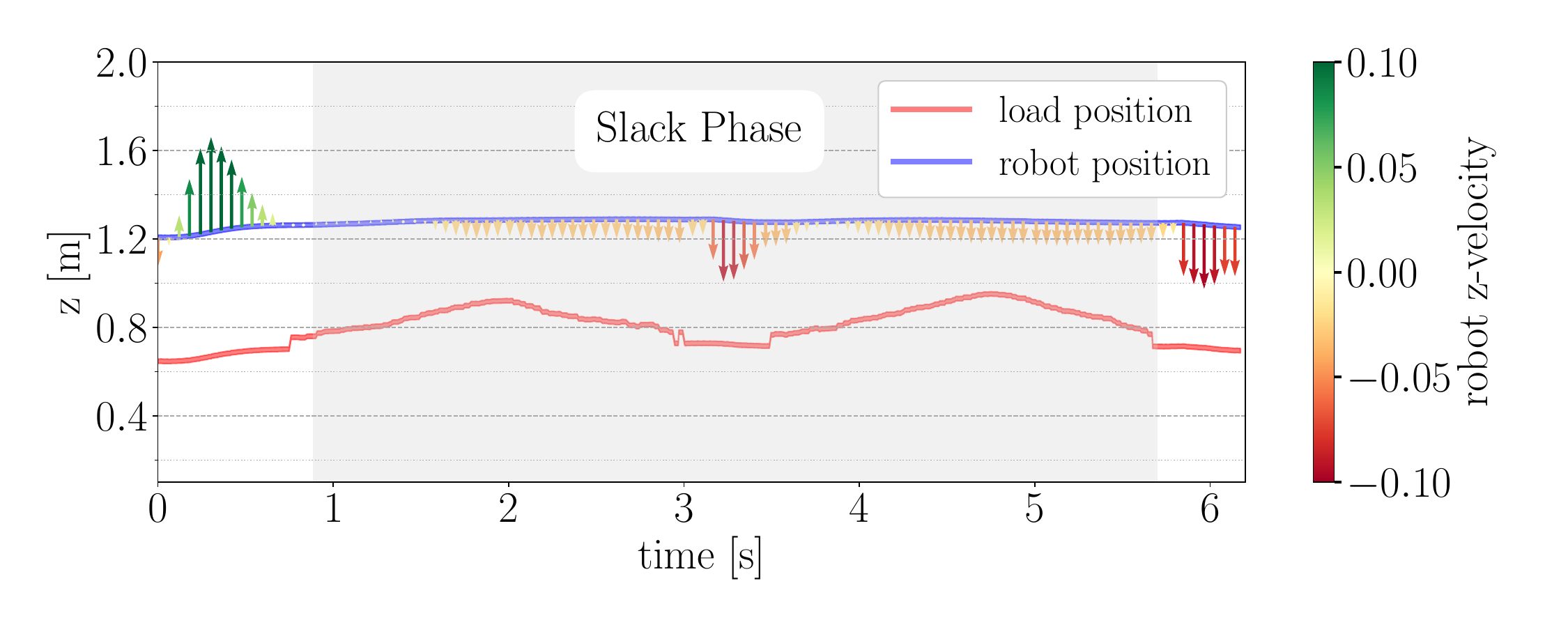}
    \end{minipage}
    \hfill
    \vspace{-15pt}
    \caption{Non-Hybrid (left) vs. Hybrid (right) controller during a slack-taut transition in a hover scenario. The hybrid controller maintains the robot's position and is unaffected by the payload movement. The non-hybrid controller incorrectly commands the robot to move towards the payload during the slack state due to $z$ error in the desired payload position. 
    % We release the payload at $t=1.2$.}% to prevent a crash.
    }
    \label{fig:hover_exp_comparison}
\end{figure*}

We show the importance of a hybrid controller during a hover scenario. We conduct this experiment by giving the robot a constant hover set-point $\vecx_{L, ref} = \begin{bmatrix} 0 ,0, 0.7\end{bmatrix}^\top$. As shown in Fig.~\ref{fig:hover_exp_comparison}, we use a gripper to manipulate the payload and displace the payload vertically to move the system into its slack phase. The differences between the two controllers can be seen by in Fig. \ref{fig:hover_exp_comparison}. When the non-hybrid controller observes a negative error in the $z$ component of $\vecx_{L}$, it assumes the cable is taut and commands the robot to move downwards to correct the error. However, this approach is ineffective because the cable is actually slack. To prevent a crash, we lift the payload up, and once the robot is close to the payload we release it, allowing the cable to become taut and letting the quadrotor regain control.
On the other hand, upon detecting the slack state, the hybrid controller transitions to using the slack state dynamics model and tracks the nominal quadrotor position using the modified HPA-MPC cost. We freely move the payload during the slack state since the robot continues to hover in place. 

\begin{figure*}[htbp]
    \begin{minipage}[b]{\columnwidth}
        \centering
        \includegraphics[width=\columnwidth, trim=0 0 60  0, clip]{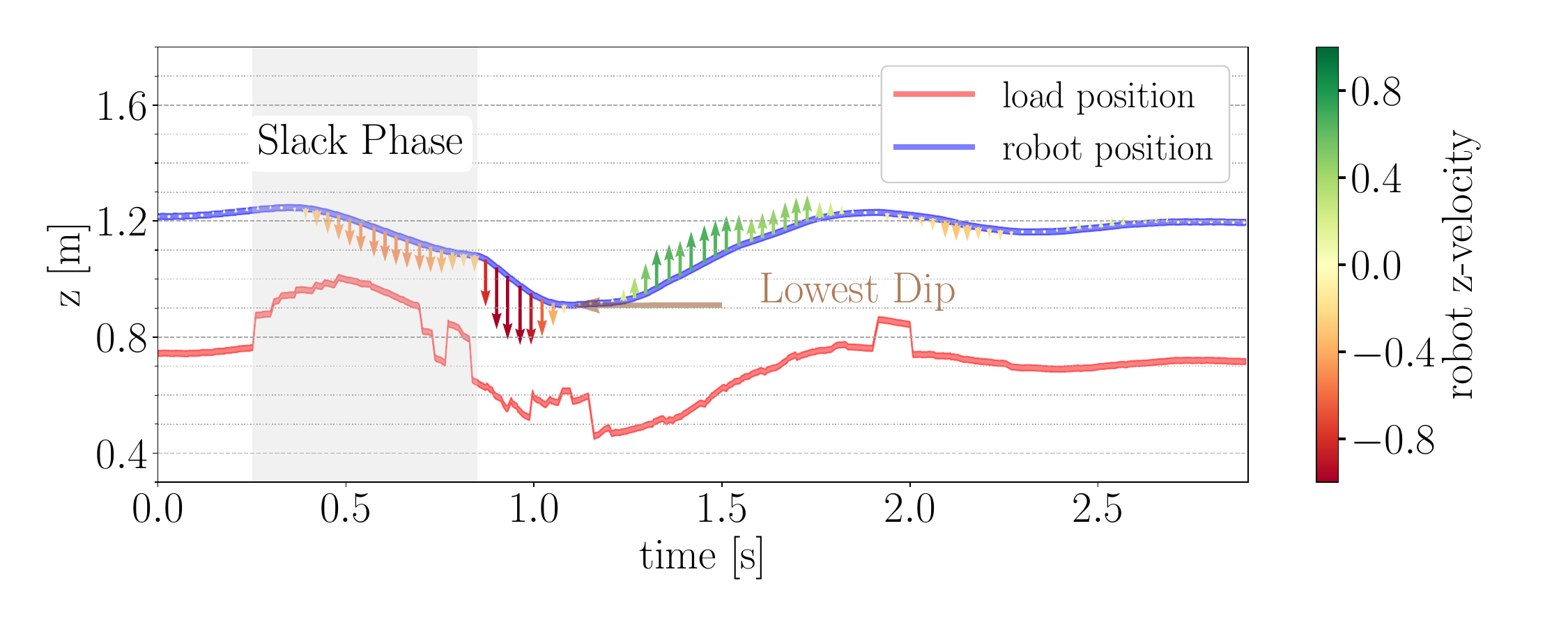}
    \end{minipage}
    \hfill
    \begin{minipage}[b]{\columnwidth}
        \centering
        \includegraphics[width=\columnwidth, trim=0 0 60 0, clip]{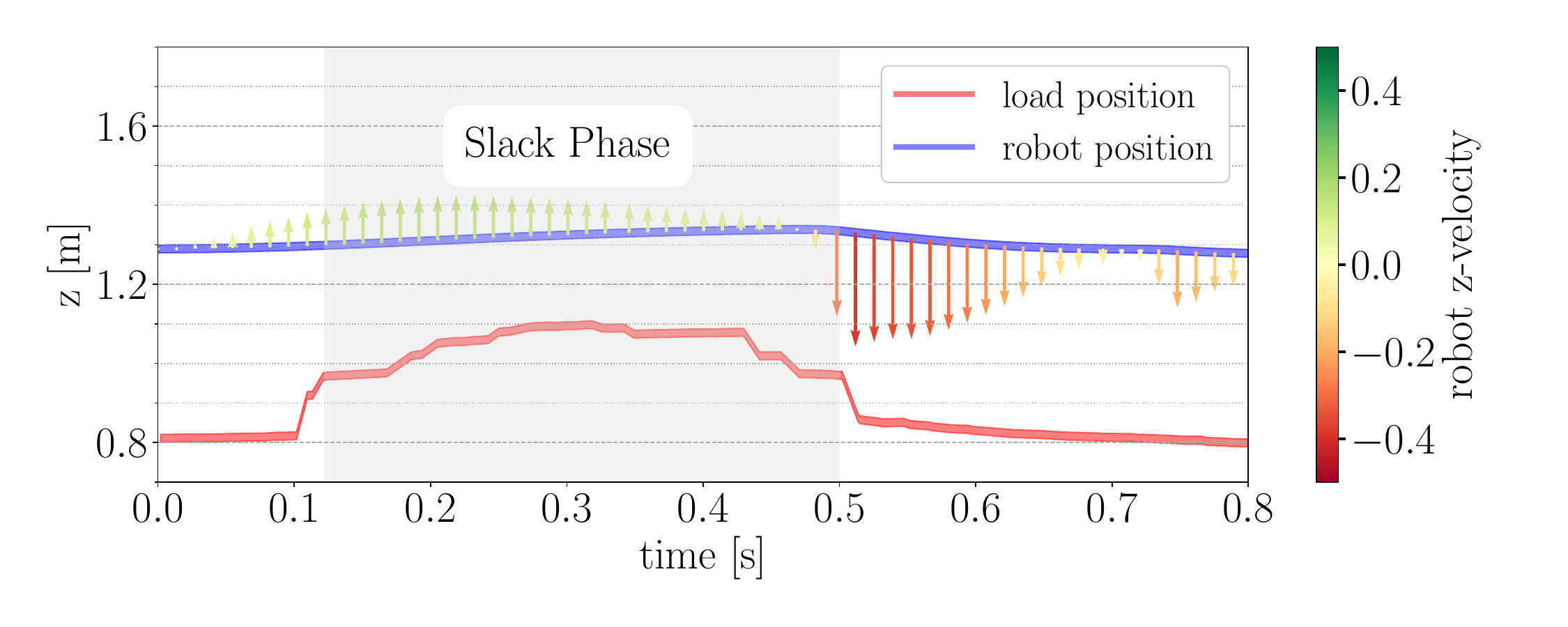}
    \end{minipage}
    \begin{minipage}[b]{\columnwidth}
        \vspace{-5pt}
        \includegraphics[width=\columnwidth, trim=0 0 0 20, clip]{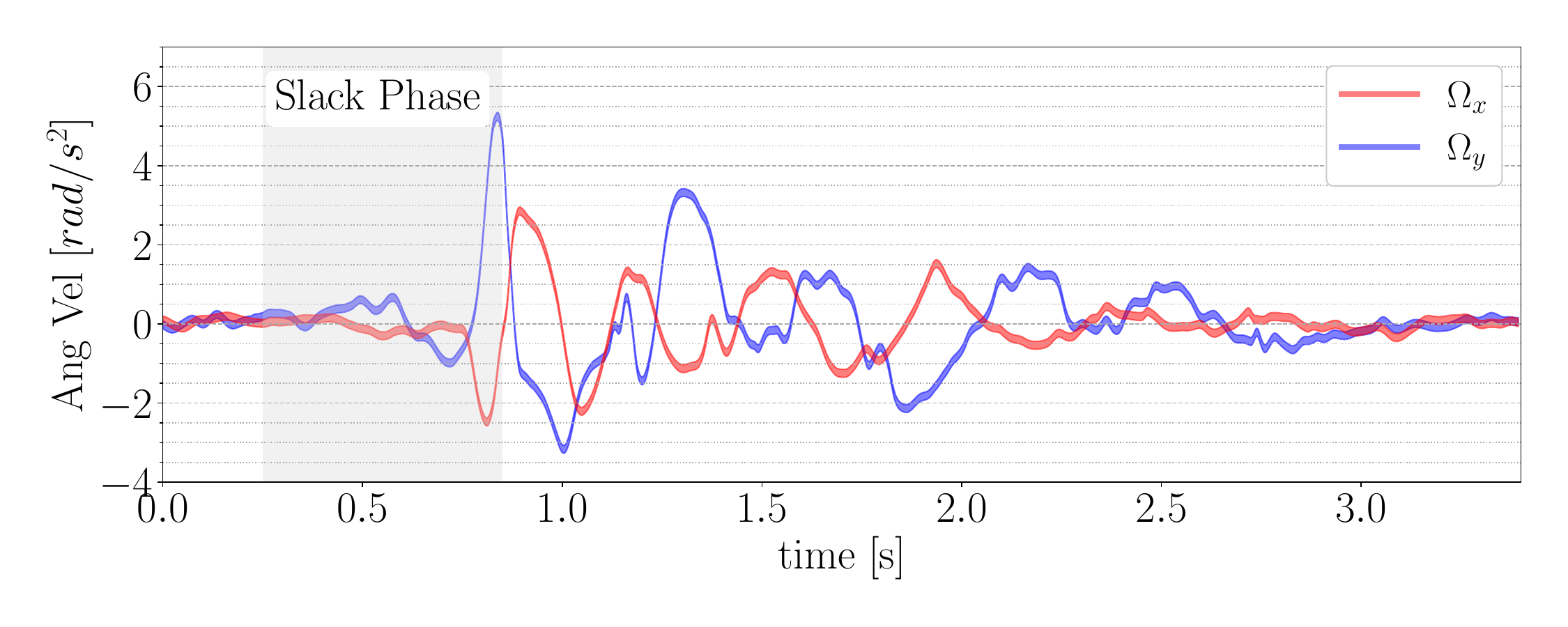}
    \end{minipage}
    \hfill
    \begin{minipage}[b]{\columnwidth}
        \vspace{-5pt}
        \includegraphics[width=\columnwidth, trim=0 0 0 20, clip]{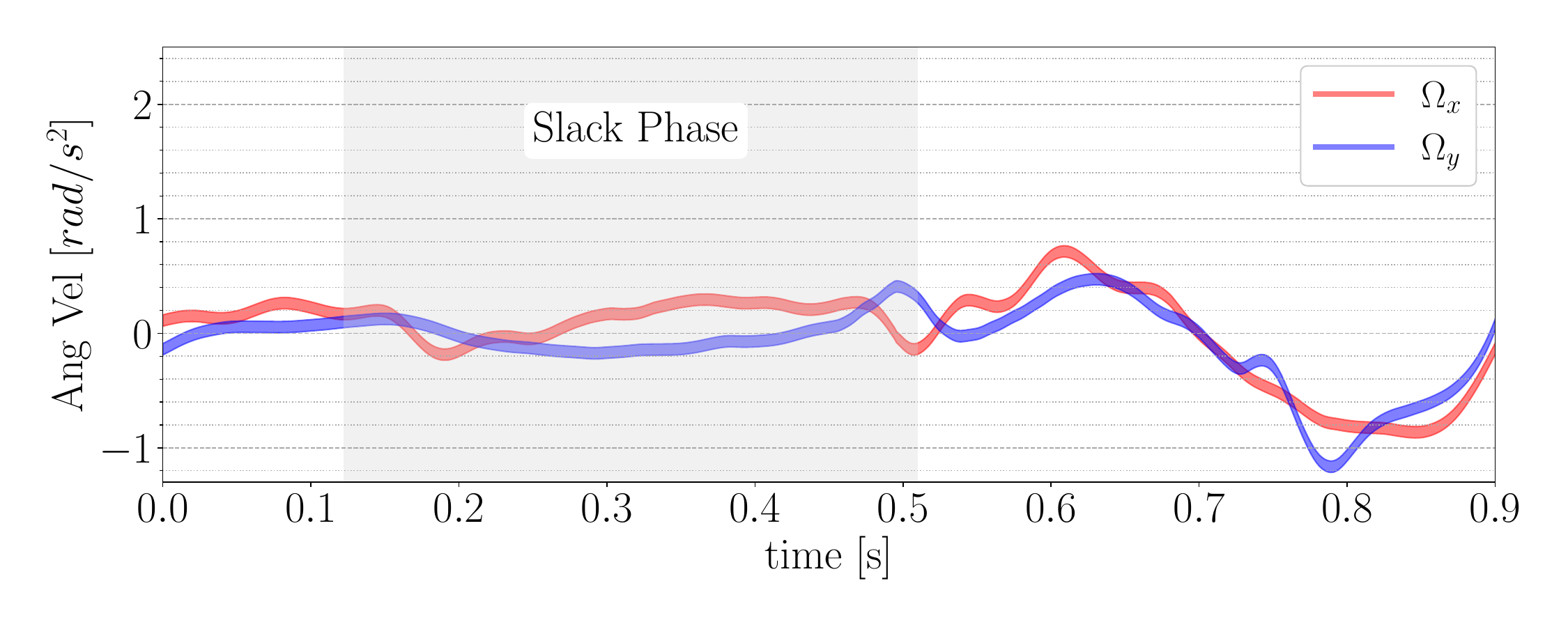}
    \end{minipage}
    \caption{Non-Hybrid (left) vs Hybrid (right) controller during a slack-taut transition during a straight line trajectory. The non-hybrid controller moves towards the payload during the slack phase. Due to this downward motion, the robot experiences a higher negative $z$ velocity and higher angular rates upon impact when compared to our hybrid controller. The hybrid controller commands the robot to track the nominal quadrotor trajectory during the slack phase and stably continues flight once the cable is taut.}
    \label{fig:traj_exp_gif}
    \vspace{-10pt}
\end{figure*}

Our method effectively handles the step change in dynamics during a mode transition by minimizing the discontinuities in control actions by leveraging the RTI scheme and a high control frequency. Due to the RTI warm start procedure, the solver is biased towards staying close to the previous commanded actions, reducing the discontinuity of the control actions. Furthermore, the high control frequency of $150$ Hz ensures rapid feedback, allowing the system to quickly compensate for unexpected deviations and achieve smoother state transitions as shown in Fig. \ref{fig:hover_exp_comparison}-\ref{fig:traj_exp_gif}.

\subsubsection{Trajectory Tracking} \label{subsec:traj_exp} 
To further validate our controller, we conduct a dynamic experiment involving the application of a vertical impulse force to the payload during the execution of a straight line trajectory. Snapshots of the experiment are shown in Fig. \ref{fig:traj_exp_gif}. During trajectory execution, we introduce a random vertical disturbance to push the payload up, causing the cable to go slack. The differences between the two controllers are shown in Fig. \ref{fig:traj_exp_gif}. Similar to the previous hover scenario, the non-hybrid controller commands undesirable actions upon detecting the cable slack state. When an error in the $z$ component of $\vecx_{L}$ is observed, the controller commands the robot to move downwards. As the payload transitions back to its taut state, it exerts an impact force onto the downward accelerating quadrotor, resulting in a substantial negative $z$ velocity and mild instability. The corresponding angular velocities are shown in Fig. \ref{fig:traj_exp_gif}. We highlight that this instability is not observed with the HPA-MPC, which maintains the robot's height during the slack phase. Although a negative $z$ velocity was still present when the cable became taut, it was significantly lower compared to the non-hybrid scenario.

\subsection{Payload Manipulation with Perception Awareness} \label{subsec:pampc_exp}
We conduct an experiment to demonstrate the capability of incorporating alternate objectives into HPA-MPC. Readers are encouraged to envision a construction scenario where a drone assists a human to install equipment. While the drone hovers, the human manipulates the object, causing the cable to go slack. During this maneuver, it is essential for the payload to remain within the camera's FoV. This ensures that if the system moves to its taut state, the robot can regain control over the payload and seamlessly maintain its position. To achieve this behavior, we add the perception aware cost described in eq.~\eqref{eq:pampc_cost}.

% \begin{figure}[htbp]
%     \begin{minipage}[b]{\columnwidth}
%          \includegraphics[width=\columnwidth, trim=40 20 20 0, clip]{figures/plots_larger_font_v2/july_1_7_perception_aware_moves_combined.pdf}
%     \end{minipage}
%     \caption{Robot x (left) and y (right) positions during a payload manipulation task. The user manipulates the payload while the HPA-MPC reacts to leep the load in its FoV.}
%     \label{fig:pampc_exp_fig}
% \end{figure}
\begin{figure}[htbp]
    \begin{minipage}[b]{\columnwidth}
    \includegraphics[width=\columnwidth, trim=65 20 5 0, clip]{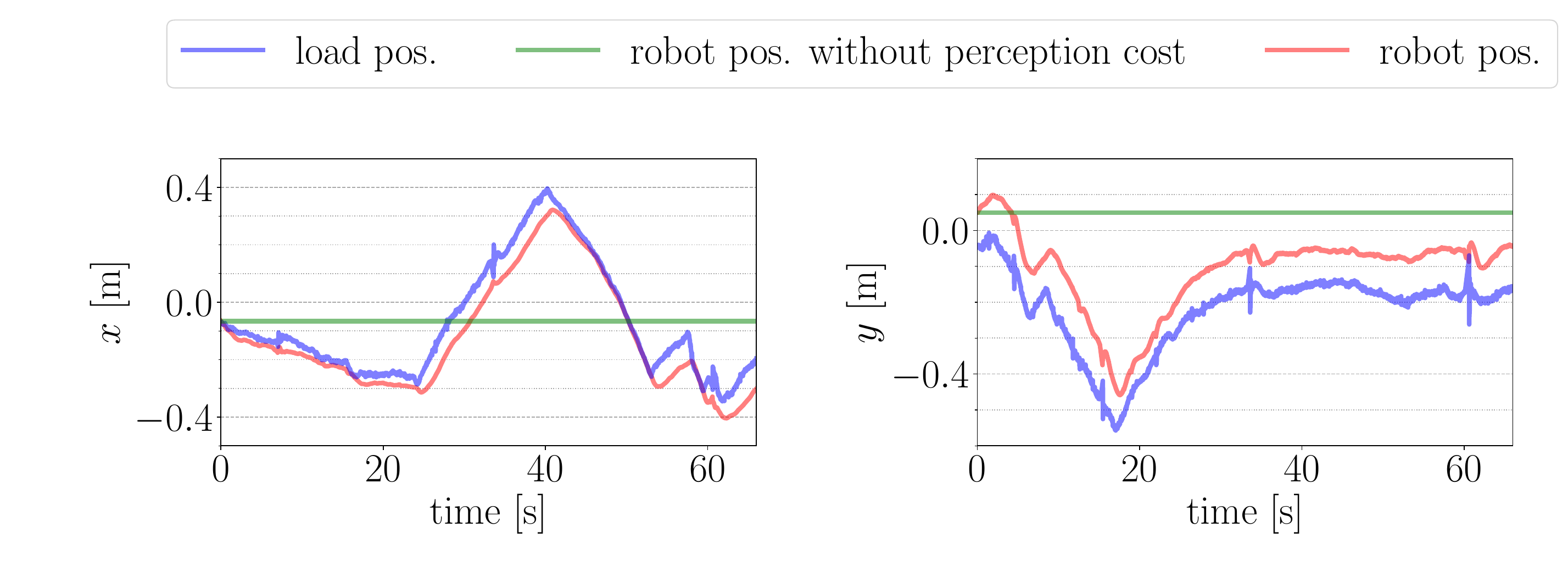}
    \end{minipage}
    \caption{Robot $x$ (left) and $y$ (right) positions during a payload manipulation task. The user manipulates the payload while the HPA-MPC reacts to leep the load in its FoV.}
    \label{fig:pampc_exp_fig}
    \vspace{-10pt}
\end{figure}

In this experiment, we provide the HPA-MPC a constant hover set point. While the robot is hovering, we use a gripper to pickup the payload and move it about the $x-y$ plane. The quadrotor responds to the payload's movement and approximately tracks it's payload's $x-y$ coordinates. The $z$ deviation of the payload did not have an effect, as the perception costs were only applied to the ${}^{\cameraf} x_{\loadf}$ and ${}^{\cameraf} y_{\loadf}$. The results for this experiment are shown in Fig.~\ref{fig:pampc_exp_fig} where the quadrotor successfully keeps the payload in it's FoV.

\section{Conclusion} \label{sec:conclusion}
In this paper, we proposed a novel HPA-MPC design for quadrotors with suspended loads and addressed the challenges of solving this problem using onboard sensing and computation. We demonstrated that the primary advantages of our approach are (i) handling unexpected slack-taut transitions induced by external disturbances while using onboard sensors to estimate the hybrid mode, (ii) maintaining the payload visibility through perception awareness during human manipulation scenarios, and (iii) successful experimental validation on small-scale aerial robots with onboard sensors and computation, despite SWaP limitations.

A few changes can be made to improve performance. First, the detected payload position and velocity can be noisy due to motion blur, which could be reduced by using traditional or learning-based motion deblurring techniques~\cite{HUIHUI2023e17332}. Next, rather than using a quadratic cost to ensure payload visibility, our future work will investigate using Control Barrier Functions to formally incorporate visibility constraints. Finally, we can further exploit our framework by allowing HPA-MPC to predict the hybrid mode while solving the OCP. This would involve solving a Mixed Integer Quadratic Program, which is beyond the scope of this work.

%\addtolength{\textheight}{-12cm} 

\bibliographystyle{IEEEtran}
\bibliography{main}

\end{document}